\documentclass[a4paper,twoside]{article}

\usepackage{natbib}
\usepackage{dblfloatfix}
\usepackage{epsfig}
\usepackage{subcaption}
\usepackage{calc}
\usepackage{amssymb}
\usepackage{amstext}
\usepackage{amsmath}
\usepackage{amsthm}
\usepackage{multicol}
\usepackage{pslatex}
\usepackage{apalike}
\usepackage{algorithm}
\usepackage{algpseudocode}
\usepackage{graphicx}
\usepackage{textcomp}
\usepackage{tabularray}
\usepackage{hhline}
\usepackage{multirow}
\usepackage{colortbl}
\usepackage{adjustbox}
\usepackage{xcolor}

\usepackage[bottom]{footmisc}
\usepackage{SCITEPRESS}     

\begin{document}

\title{Efficient Parameter Mining and Freezing for Continual Object Detection}

\author{\authorname{Angelo G. Menezes\sup{1}\orcidAuthor{0000-0002-7995-096X}, Augusto J. Peterlevitz\sup{2}\orcidAuthor{0000-0003-0575-9633}, Mateus A. Chinelatto\sup{2}\orcidAuthor{0000-0002-6933-213X}, André C. P. L. F. de Carvalho\sup{1}\orcidAuthor{0000-0002-4765-6459}}
\affiliation{\sup{1}Institute of Mathematics and Computer Sciences, University of São Paulo, São Carlos, Brazil}
\affiliation{\sup{2}Computer Vision Department, Eldorado Research Institute, Campinas, Brazil}
\email{angelomenezes@usp.br}
}

\keywords{Object Detection, Continual Learning, Continual Object Detection, Replay, Parameter Mining}

\abstract{
Continual Object Detection is essential for enabling intelligent agents to interact proactively with humans in real-world settings. While parameter-isolation strategies have been extensively explored in the context of continual learning for classification, they have yet to be fully harnessed for incremental object detection scenarios. Drawing inspiration from prior research that focused on mining individual neuron responses and integrating insights from recent developments in neural pruning, we proposed efficient ways to identify which layers are the most important for a network to maintain the performance of a detector across sequential updates. The presented findings highlight the substantial advantages of layer-level parameter isolation in facilitating incremental learning within object detection models, offering promising avenues for future research and application in real-world scenarios.}

\onecolumn \maketitle \normalsize \setcounter{footnote}{0} \vfill

\section{\uppercase{Introduction}}
\label{sec:introduction}

In the era of pervasive computing, computer vision has emerged as a central field of study with an array of applications across various domains, including healthcare, autonomous vehicles, robotics, and security systems \citep{wu2020recent}. For real-world computer vision applications, continual learning, or the ability to learn from a continuous stream of data and adapt to new tasks without forgetting previous ones, plays a vital role. It enables models to adapt to ever-changing environments and learn from a non-stationary distribution of data, mirroring human-like learning \citep{shaheen2021continual}. This form of learning becomes increasingly significant as the demand grows for models that can evolve and improve over time without the need to store all the data and be trained from scratch.

Within computer vision, object detection is a fundamental task aiming at identifying and locating objects of interest within an image. Historically, two-stage detectors, comprising a region proposal network followed by a classification stage, were the norm, but they often suffer from increased complexity and slower run-time \citep{zou2019object}. The emergence of one-stage detectors, which combine these stages into a unified framework, has allowed for more efficient and often more accurate detection \citep{tian2020fcos, lin2017focal}. In this context, incremental learning strategies for object detection can further complement one-stage detectors by facilitating the continuous adaptation of the model to new tasks or classes, making it highly suitable for real-world applications where the object landscape may change over time \citep{li2019rilod, ul2021incremental}.

Recent works have concluded that catastrophic forgetting is enlarged when the magnitude of the calculated gradients becomes higher for accommodating the new knowledge \citep{mirzadeh2021wide, hadsell2020embracing}. Since the new parameter values may deviate from the optimum place that was used to obtain the previous performance, the overall $mAP$ metrics can decline. Traditionally in continual learning (CL) for classification, researchers have proposed to tackle this problem directly by applying regularization schemes, often preventing important neurons from updating or artificially aligning the gradients for each task. Such techniques have shown fair results at the cost of being computationally expensive since network parameters are mostly adjusted individually \citep{kirkpatrick2017overcoming, chaudhry2018efficient}.

To account for the changes and keep the detector aligned with their previous performances, most works in continual object detection (COD) mitigate forgetting with regularization schemes based on complex knowledge distillation strategies and their combination with replay or the use of external data \citep{menezes2022continual}. However, we argue that the results presented by the solo work of \citet{li2018incremental} indicate that there is room to investigate further parameter-isolation schemes for COD. For these strategies, the most important neurons for a task are identified, and their changes are softened across learning updates to protect the knowledge from previous tasks. 

In this paper, we propose a thorough investigation of efficient ways to identify and penalize the change in weights for sequential updates of an object detector using insights from the neural pruning literature. We show that by intelligently freezing full significant layers of neurons, one might be able to alleviate catastrophic forgetting and foster a more efficient and robust detector.


\section{\uppercase{Related Work}}

The concept of using priors to identify the importance of the weights and protect them from updating is not new in CL. \citet{kirkpatrick2017overcoming} proposed a regularization term on the loss function that penalizes the update of important parameters. These parameters are estimated by calculating the Fish information matrix for each weight, which considers the distance between the current weight values and the optimal weights obtained when optimizing for the previous task. \citep{zenke2017continual} similarly regularized the new learning experiences but kept an online estimate of the importance of each parameter. Both strategies compute the change needed for each individual parameter, which can be computationally challenging for large-scale detectors.

Also, on the verge of regularization, \citet{li2017learning} saved a copy of the model after training for each task and, when learning a new task, applied knowledge distillation on the outputs to make sure the current model could keep responses close to the ones produced in previous tasks. Such a strategy was adapted for COD in the work of \citet{shmelkov2017incremental}, which proposed to distill knowledge from the final logits and bounding box coordinates. \citet{li2019rilod} went further and introduced an additional distillation on intermediate features for the network. Both strategies have been used in several subsequent works in COD as strong baselines for performance comparison.

In CL for classification, \citet{mallya2018packnet} conceptualized PackNet, which used concepts of the neural pruning literature for applying an iterative parameter isolation strategy. It first trained a model for a task and pruned the lowest magnitude parameters, as they were seen as the least contributors to the model's performance. Then, the left parameters were fine-tuned on the initial task data and kept frozen across new learning updates. Such a strategy is usually able to mitigate forgetting, through the cost of lower plasticity when learning new tasks. Similarly, \citet{li2018incremental} proposed a strategy, here denoted as MMN, to ``mine'' important neurons for the incremental learning of object detectors. Their method involved ranking the weights of each layer in the original model and retaining (i.e., fixing the value of) the Top-K neurons to preserve the discriminative information of the original classes, leaving the other parameters free to be updated but not zeroed as initially proposed by PackNet. The importance of each neuron is estimated by sorting them based on the absolute value of their weight. The authors evaluated this strategy with variations of the percentage of neurons to be frozen and found that a 75\% value was ideal for a stability-plasticity balance within the model. Although simple, the final described performance was on par with the state-of-the-art of the time \citep{shmelkov2017incremental}.

The above parameter-isolation strategies for CL consider that the most important individual neurons will present the highest absolute weight values and must be kept unchanged when learning new tasks. This is a traditional network pruning concept and is commonly treated as a strong baseline \citep{lecun1989optimal, li2016pruning}. 
However, Neural Network Pruning strategies have evolved to also consider the filter and layer-wise dynamics. For that, the importance of a filter or the whole layer can be obtained by analyzing the feature maps after the forward pass of a subset of the whole dataset. Then, they can be ranked and pruned based on criteria such as proximity to zero, variation inter samples, or information entropy \citep{liu2019channel, luo2017entropy, wang2021filter}. Even so, the available network capacity will be dependent on the number of involved tasks since important parameters are not allowed to change. 

\section{\uppercase{Methodology}}

Based on the recent neural pruning literature, we explore four different ways to identify important parameters to be kept intact across sequential updates. The following criteria are used to determine the importance of each network $layer$ after forwarding a subset of images from the task data and analyzing the generated feature maps:

\begin{figure*}[bp]
\centering
\includegraphics[width=\linewidth]{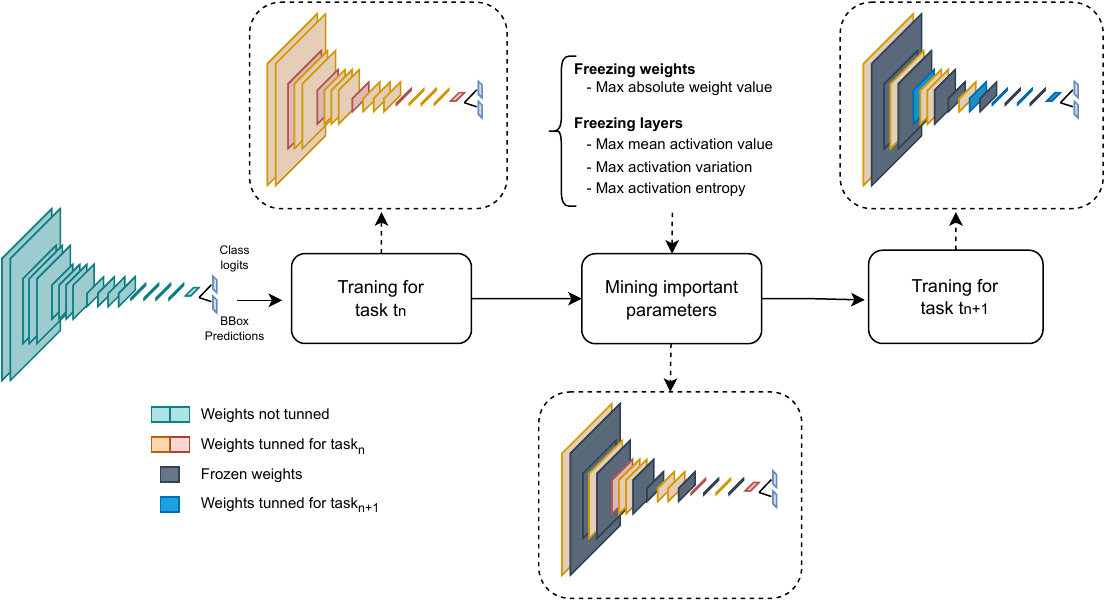}
\caption{Mining important parameters for efficient incremental updates.}
\label{fig:strategy}
\end{figure*}

\begin{itemize}
    \item \textbf{Highest mean of activation values}: Rank and select the layers with filters that produced the highest mean of activations.
    \begin{equation}
        I(layer_{i}) = \frac{1}{N}\sum_{k=1}^{N} F(x_{k}) 
    \end{equation}
    \item \textbf{Highest median of activation values}: An alternative that considers the highest median of activations instead of the mean.
    \begin{equation}
        I(layer_{i}) = Med(F(x_{k}))
    \end{equation}
    \item \textbf{Highest variance}: For this criterion, we consider that filters with higher standard deviation in the generated feature maps across diverse samples are more important and their layer should be kept unchanged.
    \begin{equation}
        I(layer_{i}) = \sqrt{\frac{1}{N} \sum_{k=1}^{N} (F(x_k) - \mu)^2}
    \end{equation}
    \item \textbf{Highest information entropy}: Rank and select the layers based on the highest information entropy on their feature maps. 
    \begin{equation}
       I(layer_{i}) = -\sum_{k=1}^{N} P(F(x_k)) \log_2 P(F(x_k))
    \end{equation}
\end{itemize}

\noindent where $N$ is the number of images in the subset; $F(x_{k})$ is the flattened feature map; $Med$ is the median of the feature map activations; $\mu$ is mean of the feature map activations; $P$ is the probability distribution of a feature map.

Additionally, in a separate investigation, we explore whether relaxing the fixed weight constraint proposed by MMN can allow the model to be more plastic while keeping decent performance on previous tasks. For that, we propose to simply adjust the changes to the mined task-specific parameters during the training step by multiplying the gradients calculated in the incremental step by a penalty value. By allowing them to adjust the important weights in a minimal way (i.e., with a penalty of 1\% or 10\%) across tasks, we hypothesize that the model will be able to circumvent capacity constraints and be more plastic. 

For the proposed layer-mining criteria, we also check which percentage (i.e., 25, 50, 75, 90) of frozen layers would give the best results. Figure \ref{fig:strategy} describes the proposed experimental pipeline.

\subsection{Evaluation Benchmarks}

Two different incremental learning scenarios were used to check the performance of the proposed methods.

\textbf{Incremental Pascal VOC}

\begin{figure}[!h]
\centering
\includegraphics[width=\linewidth]{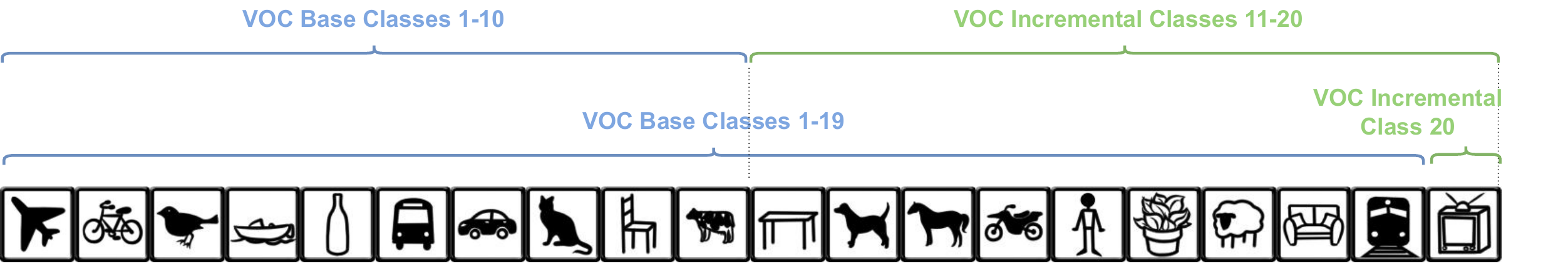}
\caption{Incremental PASCAL VOC Benchmark Evaluated Scenarios.}
\label{fig:pascal}
\end{figure}

We opted to use the incremental version of the well-known Pascal VOC dataset following the 2-step learning protocol used by the majority of works in the area \citep{menezes2022continual}. We investigated the scenarios in which the model needs to learn either the last class or the last 10 classes at once, as described in Figure \ref{fig:pascal}.

\textbf{TAESA Transmission Towers Dataset}

The detection of transmission towers and their components using aerial footage is an essential step for performing inspections on their structures. These inspections are often performed by onsite specialists to categorize the health aspect of each component. The advantage of automating such tasks by the use of drones has been largely approached in this industry setting and is known to have a positive impact on standardization of the acquisition process and reducing the number of accidents \textit{in locu}. However, there is a lack of successful reports of general applications in this field since it inherently involves several challenges related to acquiring training data, having to deal with large domain discrepancies (since energy transmission towers can be located anywhere in a country), and the necessity to update the model every time a new accessory or tower needs to be mapped.

\begin{figure}[!htbp]
\centering
\includegraphics[width=\linewidth]{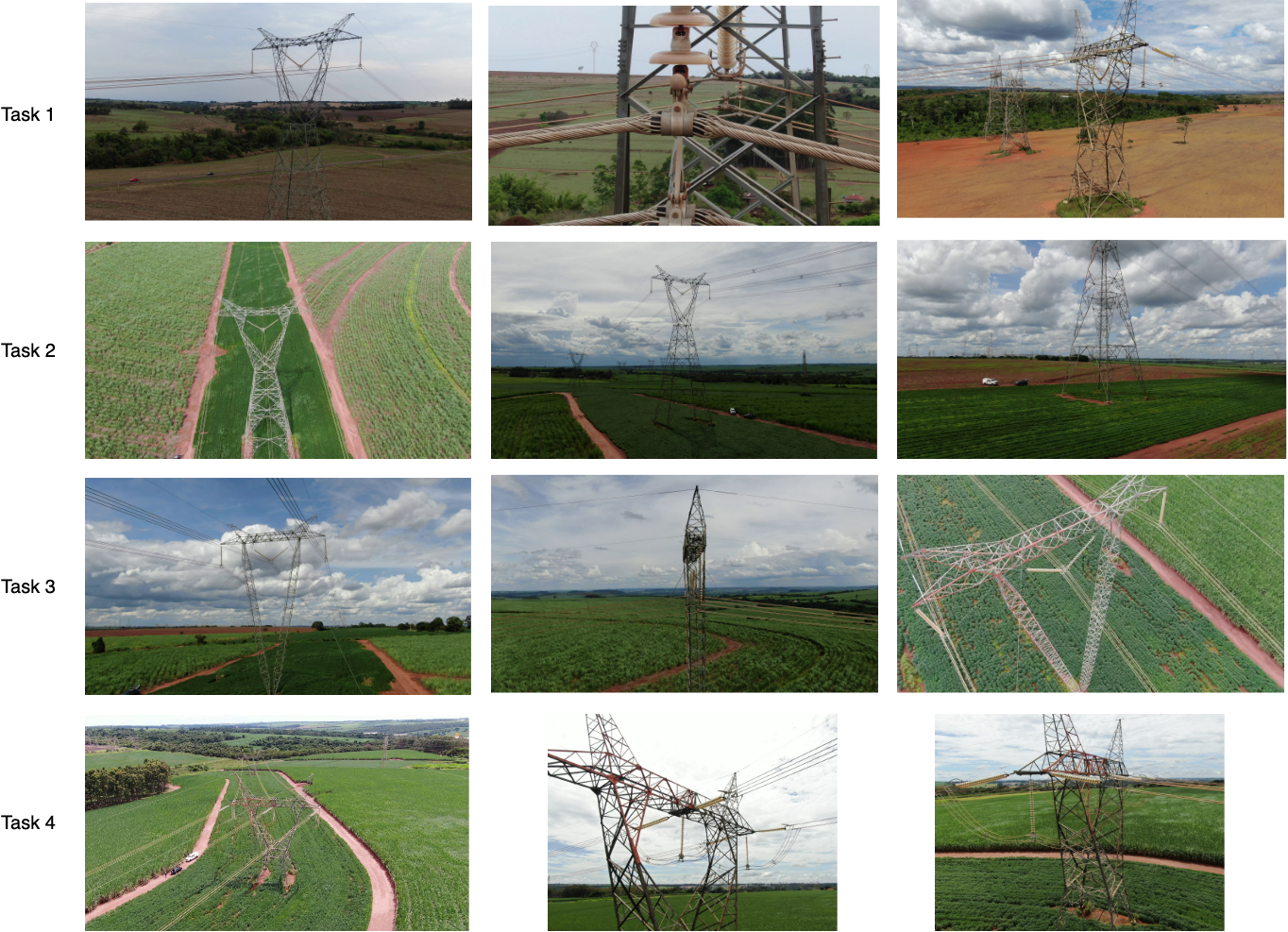}
\caption{Sample of images of each task for the TAESA Transmission Towers Dataset.}
\label{fig:taesa-sample}
\end{figure}

To aid in the proposal of solutions for some of the listed issues, we introduce the TAESA Transmission Towers Dataset. It consists of aerial images from several drone inspections performed on energy transmission sites maintained by the TAESA company in Brazil. The full dataset has records from different transmission sites from four cities with different soil and vegetation conditions. In this way, the incremental benchmark was organized into four different learning tasks, each representing data from a specific transmission site, as illustrated by Figure \ref{fig:taesa-sample}. 

\begin{table*}[!ht]
\centering
\caption{TAESA Dataset Summary.}
\label{tab:taesa-summary}
\scalebox{1}{
\begin{tabular}{llrrrrrrrrrrr} 
\cline{4-13}
                             &                         & \multicolumn{1}{l}{}                                                              & \multicolumn{10}{c}{N$^\circ$~of Boxes per label}                                                                                                                                                                                                                                                 \\ 
\hline
\multicolumn{1}{r}{Scenario} & \multicolumn{1}{c}{Set} & \multicolumn{1}{c}{\begin{tabular}[c]{@{}c@{}}N$^\circ$ of\\ Images\end{tabular}} & \multicolumn{1}{c}{0} & \multicolumn{1}{c}{1} & \multicolumn{1}{c}{2} & \multicolumn{1}{c}{3} & \multicolumn{1}{c}{4} & \multicolumn{1}{c}{5} & \multicolumn{1}{c}{6} & \multicolumn{1}{c}{7} & \multicolumn{1}{c}{8} & \multicolumn{1}{c}{\begin{tabular}[c]{@{}c@{}}Total\\Boxes\end{tabular}}  \\ 
\hline
\multirow{3}{*}{Task 1}      & Training                & 526                                                                               & 690                   & 2228                  & 482                   & 119                   & 381                   & 528                   & -                     & -                     & -                     & 4428                                                                      \\
                             & Validation              & 67                                                                                & 78                    & 245                   & 55                    & 16                    & 29                    & 49                    & -                     & -                     & -                     & 472                                                                       \\
                             & Testing                 & 69                                                                                & 91                    & 252                   & 49                    & 10                    & 42                    & 60                    & -                     & -                     & -                     & 504                                                                       \\ 
\hline
\multirow{3}{*}{Task 2}      & Training                & 431                                                                               & 86                    & 950                   & 260                   & 4                     & -                     & -                     & 20                    & 429                   & 8                     & 1757                                                                      \\
                             & Validation              & 55                                                                                & 14                    & 120                   & 32                    & -                     & -                     & -                     & 2                     & 55                    & -                     & 223                                                                       \\
                             & Testing                 & 55                                                                                & 2                     & 120                   & 29                    & 1                     & -                     & -                     & 3                     & 55                    & -                     & 210                                                                       \\ 
\hline
\multirow{3}{*}{Task 3}      & Training                & 308                                                                               & 5                     & 726                   & 269                   & 39                    & -                     & -                     & 303                   & -                     & 4                     & 1346                                                                      \\
                             & Validation              & 39                                                                                & 3                     & 92                    & 31                    & 5                     & -                     & -                     & 36                    & -                     & -                     & 167                                                                       \\
                             & Testing                 & 39                                                                                & 1                     & 89                    & 33                    & 6                     & -                     & -                     & 38                    & -                     & -                     & 167                                                                       \\ 
\hline
\multirow{3}{*}{Task 4}      & Training                & 227                                                                               & 5                     & 1242                  & 357                   & -                     & 770                   & 83                    & -                     & -                     & 234                   & 2691                                                                      \\
                             & Validation              & 28                                                                                & 2                     & 165                   & 50                    & -                     & 98                    & 12                    & -                     & -                     & 29                    & 356                                                                       \\
                             & Testing                 & 29                                                                                & -                     & 177                   & 52                    & -                     & 112                   & 11                    & -                     & -                     & 29                    & 381                                                                       \\
\hline
\end{tabular}}
\end{table*}

Each task can have new classes that were not introduced before and new visuals for a previously introduced object, making it a challenging ``data-incremental'' benchmark. In addition, different from most artificial benchmarks, images were annotated by several people using a reference sheet of the possible classes that could be present. For that, the possibility of missing annotations and label conflict in posterior tasks was reduced. A summary of the dataset with respect to the number of images and objects, with their description, for each task can be seen in Tables \ref{tab:id-class-taesa} and \ref{tab:taesa-summary}.

\begin{table}[!htb]
\centering
\caption{ID for each class in the TAESA dataset.}
\label{tab:id-class-taesa}
\begin{tabular}{cc} 
\hline
Class Label & Description  \\ 
\hline
0                    & Classic Tower         \\
1                    & Insulator             \\
2                    & Yoke Plate            \\
3                    & Clamper               \\
4                    & Ball Link             \\
5                    & Anchoring Clamp       \\
6                    & Guyed Tower           \\
7                    & Support Tower         \\
8                    & Anchor Tower          \\
\hline
\end{tabular}
\end{table}

\subsection{Implementation Details}

We opted to explore the RetinaNet one-stage detector using a frozen ResNet50 with an unfrozen FPN backbone. The selected freezing criteria is therefore only applied to the neck (i.e., FPN) and head of the model. The training settings are similar to the ones proposed by \citet{shmelkov2017incremental}. For both benchmarks, the model was trained with SGD for 40k steps with an LR of 0.01 for learning the first task. For the incremental tasks, in the Pascal VOC Benchmark, the model was trained with an LR of 0.001 for more 40k steps when presented with data from several classes and for 5k steps when only data from the last class was used. For the incremental tasks with the TAESA benchmark, the model was trained with an LR of 0.001 for 5k steps for each new task. The code for training the network was written in Python and used the MMDetection toolbox for orchestrating the detection benchmark and evaluation procedure \citep{chen2019mmdetection}. The main followed steps are depicted below in Algorithm \ref{multi-task-algorithm}.

\begin{algorithm}
\caption{Incremental training with parameter mining and freezing for COD}\label{multi-task-algorithm}
\begin{algorithmic}[1]
\State M: Model to be trained
\State $Tasks$: List of learning experiences
\State $S$: Type of mining strategy
\State $L$: Percentage $L$ of frozen layers or parameters
\State $P$: Percentage of gradient penalty
\State $C$: Criteria for freezing the layers
\State $N$: Percentage of samples from $Task_{i}$ to be used for calculating freezing metrics
\State $i \gets 0$
\For{$i$ in range(length($Tasks$))}:
    \State Train model $M$ with data from $Task_{i}$
    \If {$S$ $= gradient\_mining$}
        \State Dump previous gradient hooks
        \State Attach a hook with the gradient penalty $P$ to the selected percentage $L$ of parameters 
    \EndIf
    \If {$S$  $= layer\_freezing$}
        \State Reset $requires\_grad$ of the parameters in each layer
        \State Freeze a percentage $L$ of the layers given the chosen criteria $C$ using statistics from the feature maps obtained after forwarding the $N$ selected samples
    \EndIf
    \State Fine-tune in $Task_{i}$ for $1k$ steps to regularize parameters for the next learning experience
    \State $i \gets i + 1$
\EndFor
\State \textbf{return} $M$
\end{algorithmic}
\end{algorithm}

As for the baselines, for the Incremental Pascal VOC benchmark, we considered the results reported on the work of \citet{li2019rilod} for the ILOD and RILOD strategies which also made use of the RetinaNet with ResNet50 as the backbone in a similar training setting. For the TAESA benchmark, we propose the comparison against Experience Replay using a task-balanced random reservoir buffer. We also compare the results in both benchmarks against our implementation of the MMN strategy from \citet{li2018incremental} as well as the upper bound when all data is available for training the model. To account for the randomness associated with neural networks, we report the performance of each strategy after the averaging of three runs with different seeds.

\subsection{Evaluation Metrics}

For checking the performance in the Incremental Pascal VOC benchmark, we use the average $mAP[.5]$ and $\Omega$ for comparisons against the upper bound (i.e., join-training) as usually reported by other works. To better evaluate the potential of each strategy regarding the model`s ability to retain and acquire new knowledge, we also apply the metrics proposed by \citet{menezes2022continual} known as the rate of stability ($RSD$) and plasticity ($RPD$) deficits, described in Equations~\ref{eq-rate-stab} and \ref{eq-rate-plast}.



\begin{equation}
    \label{eq-rate-stab}
    \begin{aligned}[b]
        & \text{RSD} = \frac{1}{N_{old\_classes}} \times\\
        & \sum_{i=1}^{N_{old\_classes}}\frac{mAP_{joint,i} - mAP_{inc,i}}{mAP_{joint,i}} \ * 100
    \end{aligned}
\end{equation}

\begin{equation}
    \label{eq-rate-plast}
    \begin{aligned}[b]
        & \text{RPD} = \frac{1}{N_{new\_classes}} \times\\
        & \sum_{i=N_{old\_classes}+1}^{N_{new\_classes}}\frac{mAP_{joint,i} - mAP_{inc,i}}{mAP_{joint,i}} \ * 100
    \end{aligned}
\end{equation}

Especially for the TAESA benchmark, the performance is measured by the final $mAP$, with different thresholds, and $mAP[.50]$ after learning all tasks, as well as with their upper-bound ratios $\Omega_{mAP}$ and $\Omega_{mAP[.50]}$. Additionally, since the benchmark involves the introduction of a sequence of tasks, we have modified the existing $RSD$ and $RPD$ metrics to consider individual tasks instead of classes. In this evaluation scenario, $RSD$ measures the performance deficit against the upper bound $mAP$ in all tasks up to the last one, while $RPD$ evaluates the performance deficit against the last learned task.

\section{\uppercase{Results}}
\label{sec:results}

\subsection{Pascal VOC 1-19 + 20}

\begin{table*}[!htb]
\centering
\caption{Results when learning the last class (TV monitor)}\label{tab:results-min-voc-1}
\scalebox{0.5}{
\begin{tabular}{lrrrrrrrrrrrrrrrrrrrrrrrrr} 
\hline
\multicolumn{2}{c}{\textbf{\textcolor[rgb]{0,0.502,0}{19 + 1}}}                           & \multicolumn{1}{c}{aero} & \multicolumn{1}{c}{cycle} & \multicolumn{1}{c}{bird} & \multicolumn{1}{c}{boat} & \multicolumn{1}{c}{bottle} & \multicolumn{1}{c}{bus} & \multicolumn{1}{c}{car} & \multicolumn{1}{c}{cat} & \multicolumn{1}{c}{chair} & \multicolumn{1}{c}{cow} & \multicolumn{1}{c}{table} & \multicolumn{1}{c}{dog} & \multicolumn{1}{c}{horse} & \multicolumn{1}{c}{bike} & \multicolumn{1}{c}{person} & \multicolumn{1}{c}{plant} & \multicolumn{1}{c}{sheep} & \multicolumn{1}{c}{sofa} & \multicolumn{1}{c}{train} & \multicolumn{1}{c}{{\cellcolor[rgb]{0.812,0.886,0.953}}tv} & \multicolumn{1}{c}{\textbf{mAP}} & \multicolumn{1}{c}{$\Omega_{all} \uparrow$} & \multicolumn{1}{c}{\textbf{RSD }($\%$)$\downarrow$} & \multicolumn{1}{c}{\textbf{RPD} ($\%$)$ \downarrow$}  \\ 
\hline
\multicolumn{2}{l}{Upper-bound}                                                           & 73.5                     & 80.6                      & 77.4                     & 61.2                     & 62.2                       & 79.9                    & 83.4                    & 86.7                    & 47.6                      & 78                      & 68.1                      & 85.1                    & 83.7                      & 82.8                     & 79.1                       & 42.5                      & 75.7                      & 64.9                     & 79                        & {\cellcolor[rgb]{0.812,0.886,0.953}}76.2                   & 73.4                             & -                                           & -                                                   & -                                                     \\
\multicolumn{2}{l}{First 19}                                                              & 77                       & 83.5                      & 77.7                     & 65.1                     & 63                         & 78.1                    & 83.6                    & 88.5                    & 55.2                      & 79.7                    & 71.3                      & 85.8                    & 85.2                      & 83                       & 80.2                       & 44.1                      & 75.2                      & 69.7                     & 81.4                      & {\cellcolor[rgb]{0.812,0.886,0.953}}0                      & 71.4                             & -                                           & -                                                   & -                                                     \\
\multicolumn{2}{l}{New 1}                                                                 & 48                       & 61.2                      & 27.6                     & 18.1                     & 8.1                        & 58.7                    & 53.4                    & 17.1                    & 0                         & 45.9                    & 18.2                      & 31.9                    & 59.9                      & 62.2                     & 9.1                        & 3.4                       & 42.9                      & 0                        & 50.3                      & {\cellcolor[rgb]{0.812,0.886,0.953}}63.8                   & 34.0                             & -                                           & -                                                   & -                                                     \\ 
\hline
\multicolumn{2}{l}{ILOD}                                                                  & 61.9                     & 78.5                      & 62.5                     & 39.2                     & 60.9                       & 53.2                    & 79.3                    & 84.5                    & 52.3                      & 52.6                    & 62.8                      & 71.5                    & 51.8                      & 61.5                     & 76.8                       & 43.8                      & 43.8                      & 69.7                     & 52.9                      & {\cellcolor[rgb]{0.812,0.886,0.953}}44.6                   & 60.2                             & 0.81                                        & 18.01                                               & 45.66                                                 \\
\multicolumn{2}{l}{RILOD}                                                                 & 69.7                     & 78.3                      & 70.2                     & 46.4                     & 59.5                       & 69.3                    & 79.7                    & 79.9                    & 52.7                      & 69.8                    & 57.4                      & 75.8                    & 69.1                      & 69.8                     & 76.4                       & 43.2                      & 68.5                      & 70.9                     & 53.7                      & {\cellcolor[rgb]{0.812,0.886,0.953}}40.4                   & 65.0                             & 0.87                                        & 10.90                                               & 51.28                                                 \\ 
\hline
\multirow{4}{*}{MMN}                                                                 & 25 & 71.8                     & 78.8                      & 66.5                     & 48.5                     & 48.6                       & 73.4                    & 78.8                    & 77.1                    & 9.1                       & 76.5                    & 52.3                      & 74.7                    & 82.4                      & 76.3                     & 62.3                       & 21.5                      & 65.9                      & 20.9                     & 68.2                      & {\cellcolor[rgb]{0.812,0.886,0.953}}45.6                   & 60.0                             & 0.82                                        & 17.06                                               & 41.70                                                 \\
                                                                                     & 50 & 73.4                     & 79                        & 71.5                     & 51                       & 53.4                       & 73.4                    & 81.6                    & 78.5                    & 13.9                      & 73.5                    & 54.5                      & 76.7                    & 83.2                      & 79.1                     & 64                         & 27.7                      & 66.8                      & 36.3                     & 69.4                      & {\cellcolor[rgb]{0.812,0.886,0.953}}43                     & 62.5                             & 0.85                                        & 13.23                                               & 45.24                                                 \\
                                                                                     & 75 & 74.8                     & 79.3                      & 72.9                     & 54.9                     & 54                         & 73.9                    & 82                      & 85                      & 25.4                      & 77.2                    & 60                        & 81.8                    & 83.5                      & 80.2                     & 70.1                       & 35.9                      & 68                        & 49.7                     & 67.8                      & {\cellcolor[rgb]{0.812,0.886,0.953}}39.3                   & 65.8                             & \textbf{0.90}                               & \textbf{8.25}                                       & \textbf{50.29}                                        \\
                                                                                     & 90 & 76.5                     & 82.4                      & 74.4                     & 58.4                     & 57.9                       & 74.2                    & 82.3                    & 86.7                    & 35.7                      & 77.6                    & 65.1                      & 83.7                    & 83.8                      & 82.2                     & 72.5                       & 37                        & 73.2                      & 58.5                     & 71.5                      & {\cellcolor[rgb]{0.812,0.886,0.953}}33.7                   & 68.4                             & 0.93                                        & 4.15                                                & 57.92                                                 \\ 
\hline\hline
\multirow{4}{*}{\begin{tabular}[c]{@{}l@{}}Gradient\\penalty\\of 1\%\end{tabular}}   & 25 & 71.9                     & 78.8                      & 66.5                     & 48.6                     & 48.5                       & 73.4                    & 78.8                    & 77.1                    & 9.1                       & 76.5                    & 52.3                      & 74.6                    & 82.4                      & 76.3                     & 62.3                       & 21.5                      & 65.9                      & 20.7                     & 68                        & {\cellcolor[rgb]{0.812,0.886,0.953}}45.5                   & 59.9                             & 0.82                                        & 17.08                                               & 41.84                                                 \\
                                                                                     & 50 & 73.3                     & 79                        & 71.4                     & 51                       & 53.3                       & 73.4                    & 81.6                    & 78.4                    & 13.8                      & 73.5                    & 54.4                      & 76.7                    & 83.2                      & 79                       & 64                         & 27.4                      & 66.8                      & 34.7                     & 69.3                      & {\cellcolor[rgb]{0.812,0.886,0.953}}43                     & 62.4                             & 0.85                                        & 13.43                                               & 45.24                                                 \\
                                                                                     & 75 & 75                       & 79.3                      & 72.9                     & 54.9                     & 54                         & 73.8                    & 82                      & 84.9                    & 25.3                      & 77.2                    & 59.8                      & 81.8                    & 83.5                      & 80.1                     & 70.1                       & 35.8                      & 67.9                      & 49.3                     & 67.8                      & {\cellcolor[rgb]{0.812,0.886,0.953}}39.4                   & 65.7                             & \textbf{0.90}                               & \textbf{8.32}                                       & \textbf{50.15}                                        \\
                                                                                     & 90 & 76                       & 82.1                      & 74.4                     & 57.3                     & 57.3                       & 74.1                    & 82.1                    & 85.9                    & 34                        & 77.4                    & 63.4                      & 82.9                    & 83.4                      & 82                       & 72.1                       & 37.1                      & 72.4                      & 57.1                     & 70.5                      & {\cellcolor[rgb]{0.812,0.886,0.953}}34.3                   & 67.8                             & 0.92                                        & 5.01                                                & 57.10                                                 \\ 
\hline
\multirow{4}{*}{\begin{tabular}[c]{@{}l@{}}Gradient\\penalty\\of 10\%\end{tabular}}  & 25 & 71.8                     & 78.6                      & 66.5                     & 48                       & 48.5                       & 73.4                    & 78.8                    & 77.1                    & 9.1                       & 76.5                    & 52.2                      & 74.1                    & 82.4                      & 76.2                     & 62.2                       & 21                        & 65.6                      & 19.9                     & 68.2                      & {\cellcolor[rgb]{0.812,0.886,0.953}}45.4                   & 59.8                             & 0.81                                        & 17.31                                               & 41.97                                                 \\
                                                                                     & 50 & 73.1                     & 78.8                      & 71.3                     & 49.6                     & 53.3                       & 74.5                    & 81.5                    & 78.3                    & 11.4                      & 73.4                    & 54                        & 76.4                    & 82.8                      & 76.8                     & 63.8                       & 27                        & 66.4                      & 33.4                     & 68.6                      & {\cellcolor[rgb]{0.812,0.886,0.953}}43.8                   & 61.9                             & 0.84                                        & 14.13                                               & 44.15                                                 \\
                                                                                     & 75 & 73.9                     & 79.2                      & 72.9                     & 53.5                     & 54.2                       & 73.4                    & 81.8                    & 79.6                    & 22                        & 76.9                    & 58.4                      & 81.6                    & 83.3                      & 79.8                     & 69.3                       & 33.6                      & 67.4                      & 47.2                     & 67.4                      & {\cellcolor[rgb]{0.812,0.886,0.953}}39.4                   & 64.7                             & 0.88                                        & 9.75                                                & 50.15                                                 \\
                                                                                     & 90 & 76.2                     & 81.8                      & 73.6                     & 55.9                     & 57                         & 73.2                    & 81.2                    & 84.6                    & 30.3                      & 76.9                    & 60.7                      & 82.4                    & 83.6                      & 81.1                     & 71.1                       & 36.3                      & 68.3                      & 56                       & 67                        & {\cellcolor[rgb]{0.812,0.886,0.953}}37.2                   & 66.7                             & \textbf{0.91}                               & \textbf{6.76}                                       & \textbf{53.15}                                        \\ 
\hline\hline
\multirow{4}{*}{\begin{tabular}[c]{@{}l@{}}Freezing\\based on\\mean\end{tabular}}    & 25 & 75.1                     & 78.8                      & 71.6                     & 57.3                     & 54.3                       & 75.3                    & 81.1                    & 78.6                    & 27.5                      & 77                      & 60.4                      & 80.8                    & 82.5                      & 79.6                     & 70.5                       & 32.5                      & 72.3                      & 57.3                     & 74.1                      & {\cellcolor[rgb]{0.812,0.886,0.953}}31.3                   & 65.9                             & 0.90                                        & 7.52                                                & 61.19                                                 \\
                                                                                     & 50 & 75.3                     & 78.6                      & 72                       & 57.7                     & 53.8                       & 74.7                    & 81                      & 79                      & 27                        & 74.7                    & 62.5                      & 77.8                    & 82.7                      & 77.5                     & 70.5                       & 33.1                      & 72                        & 56.5                     & 73.1                      & {\cellcolor[rgb]{0.812,0.886,0.953}}32.4                   & 65.6                             & 0.89                                        & 8.03                                                & 59.69                                                 \\
                                                                                     & 75 & 76                       & 79.5                      & 73.2                     & 58                       & 57                         & 75.8                    & 81.6                    & 84.4                    & 27.3                      & 77.3                    & 64.8                      & 82.1                    & 82.7                      & 80.4                     & 71.5                       & 36                        & 72.7                      & 57.4                     & 74.8                      & {\cellcolor[rgb]{0.812,0.886,0.953}}25.2                   & 66.9                             & 0.91                                        & 5.66                                                & 69.50                                                 \\
                                                                                     & 90 & 76.2                     & 81.3                      & 71.9                     & 60.8                     & 49.9                       & 75.7                    & 82.8                    & 86.2                    & 24.8                      & 76.5                    & 69.4                      & 82                      & 82.9                      & 80.9                     & 68.5                       & 26.2                      & 71.9                      & 60.3                     & 79.4                      & {\cellcolor[rgb]{0.812,0.886,0.953}}41.7                   & 67.5                             & \textbf{0.92}                               & \textbf{6.01}                                       & \textbf{47.02}                                        \\ 
\hline
\multirow{4}{*}{\begin{tabular}[c]{@{}l@{}}Freezing\\based on\\median\end{tabular}}  & 25 & 75.1                     & 78.7                      & 71.7                     & 57.3                     & 54.4                       & 74.8                    & 81.2                    & 78.7                    & 27.4                      & 76.9                    & 60.1                      & 80.8                    & 82.5                      & 79.3                     & 70.6                       & 32.3                      & 72.5                      & 57.3                     & 73.6                      & {\cellcolor[rgb]{0.812,0.886,0.953}}31.3                   & 65.8                             & 0.90                                        & 7.62                                                & 61.19                                                 \\
                                                                                     & 50 & 75.3                     & 78.8                      & 72.3                     & 57.7                     & 56.7                       & 74                      & 81.6                    & 79.4                    & 26.5                      & 76.9                    & 63.1                      & 81.8                    & 82.6                      & 78.9                     & 70.8                       & 34.7                      & 72.8                      & 56.2                     & 72.9                      & {\cellcolor[rgb]{0.812,0.886,0.953}}24.4                   & 65.9                             & 0.90                                        & 7.06                                                & 70.59                                                 \\
                                                                                     & 75 & 78                       & 79.6                      & 73.2                     & 57.1                     & 55.7                       & 76.1                    & 82.6                    & 86.1                    & 38.3                      & 77.2                    & 65.8                      & 83.1                    & 82.4                      & 80.5                     & 73.7                       & 38.5                      & 71.6                      & 60.5                     & 75.4                      & {\cellcolor[rgb]{0.812,0.886,0.953}}31.2                   & 68.3                             & 0.93                                        & 4.02                                                & 61.32                                                 \\
                                                                                     & 90 & 77.4                     & 82.1                      & 72.7                     & 61.3                     & 50.3                       & 77.2                    & 82.9                    & 85.8                    & 28.8                      & 76.4                    & 69.5                      & 82                      & 82.8                      & 81.2                     & 68.5                       & 27.5                      & 71.7                      & 60.4                     & 79.1                      & {\cellcolor[rgb]{0.812,0.886,0.953}}39.6                   & 67.9                             & \textbf{0.92}                               & \textbf{5.29}                                       & \textbf{49.88}                                        \\ 
\hline
\multirow{4}{*}{\begin{tabular}[c]{@{}l@{}}Freezing\\based on\\std\end{tabular}}     & 25 & 75.1                     & 78.9                      & 71.6                     & 57.3                     & 54.3                       & 75.3                    & 81.1                    & 78.6                    & 27.5                      & 77                      & 60.4                      & 80.8                    & 82.5                      & 77.4                     & 70.5                       & 32.4                      & 72.3                      & 57.3                     & 74                        & {\cellcolor[rgb]{0.812,0.886,0.953}}31.5                   & 65.8                             & 0.90                                        & 7.68                                                & 60.92                                                 \\
                                                                                     & 50 & 75.1                     & 78.9                      & 71.6                     & 57.2                     & 54.3                       & 75.3                    & 81.1                    & 78.7                    & 27.5                      & 77                      & 60.4                      & 80.7                    & 82.5                      & 77.4                     & 70.5                       & 32.3                      & 72.3                      & 57.3                     & 74                        & {\cellcolor[rgb]{0.812,0.886,0.953}}31.4                   & 65.8                             & 0.90                                        & 7.70                                                & 61.05                                                 \\
                                                                                     & 75 & 75.7                     & 79.1                      & 72.9                     & 57.1                     & 56.4                       & 75.2                    & 81.4                    & 79.3                    & 25.2                      & 77.4                    & 61.5                      & 81.6                    & 82                        & 79.5                     & 70.6                       & 33.7                      & 72.9                      & 56.1                     & 74.5                      & {\cellcolor[rgb]{0.812,0.886,0.953}}27.9                   & 66.0                             & 0.90                                        & 7.12                                                & 65.82                                                 \\
                                                                                     & 90 & 77.6                     & 79.9                      & 73.5                     & 57.3                     & 56.6                       & 77.7                    & 82.8                    & 86.2                    & 38.2                      & 77.1                    & 65.9                      & 82.8                    & 82.5                      & 80.2                     & 73.7                       & 39                        & 72.4                      & 61.5                     & 76                        & {\cellcolor[rgb]{0.812,0.886,0.953}}31.5                   & 68.6                             & \textbf{0.94}                               & \textbf{3.62}                                       & \textbf{60.92}                                        \\ 
\hline
\multirow{4}{*}{\begin{tabular}[c]{@{}l@{}}Freezing\\based on\\entropy\end{tabular}} & 25 & 75.5                     & 79.4                      & 72.7                     & 56.2                     & 57.2                       & 74.8                    & 81.9                    & 84.7                    & 28.9                      & 77.9                    & 62                        & 81.4                    & 83.1                      & 81.1                     & 71.6                       & 35.3                      & 68.4                      & 54.7                     & 69                        & {\cellcolor[rgb]{0.812,0.886,0.953}}40.7                   & 66.8                             & 0.91                                        & 6.86                                                & 48.38                                                 \\
                                                                                     & 50 & 76.8                     & 81.6                      & 72.5                     & 57                       & 52.2                       & 74.7                    & 83.2                    & 78.3                    & 22.2                      & 73.8                    & 63.7                      & 78.1                    & 81.3                      & 80                       & 70.7                       & 25.3                      & 71                        & 45.4                     & 74.4                      & {\cellcolor[rgb]{0.812,0.886,0.953}}57                     & 66.0                             & \textbf{0.90}                               & \textbf{9.27}                                       & \textbf{26.17}                                        \\
                                                                                     & 75 & 76.9                     & 81.8                      & 71.9                     & 61.4                     & 50.4                       & 76                      & 82.7                    & 86                      & 29.5                      & 76                      & 69.6                      & 82.3                    & 82.9                      & 80.7                     & 68.6                       & 26.7                      & 72.1                      & 60.9                     & 79.6                      & {\cellcolor[rgb]{0.812,0.886,0.953}}40.5                   & 67.8                             & 0.92                                        & 5.41                                                & 48.65                                                 \\
                                                                                     & 90 & 77.4                     & 81.9                      & 72.3                     & 61.4                     & 50.2                       & 76.3                    & 82.9                    & 85.7                    & 30                        & 76                      & 69.6                      & 82.2                    & 82.5                      & 81.2                     & 68.5                       & 27.4                      & 72                        & 60.7                     & 79.4                      & {\cellcolor[rgb]{0.812,0.886,0.953}}38.2                   & 67.8                             & 0.92                                        & 5.29                                                & 51.79                                                 \\
\hline
\end{tabular}}
\end{table*}

Table \ref{tab:results-min-voc-1} describes the performance of each strategy for the $19+1$ scenario. For this scenario, we noticed that the final $mAP$ and $\Omega_{all}$ would heavily benefit models that were more stable than plastic since there was a clear imbalance in the number of represented classes (i.e., $19 \rightarrow 1$) for the incremental step. With that in mind, we analyzed the results that better balanced the decrease in $RSD$ and $RPD$ since, by splitting the deficits in performance, it is clearer to understand the ability to forget and adapt in each model. By comparing the results of the application of gradient penalty with respect to freezing the neurons with the highest magnitude (i.e., MMN in Table \ref{tab:results-min-voc-1}), we see that allowing the extra plasticity did not produce broad effects in performance. However, when 90\% of the weights were mined, the extra adjustments introduced by using 1\% of the calculated gradients allowed the model to beat MMN. Regarding the results of layer-mining, freezing based on information entropy presented a better balance in $RSD$ and $RPD$, even against more established techniques such as ILOD and RILOD. For most of the results, increasing the percentage of frozen layers gave a lower deficit in stability with the caveat of increasing the difference in $mAP$ against the upper bound for the new learned class.

Overall, leaving a lower percentage of parameters frozen across updates for the methods that worked on individual neurons made their networks more adaptable. Yet, this hyperparameter for the layer-freezing methods did not greatly affect the learning of the new class but had a significant impact on the detection of classes that had been learned previously.

\subsection{Pascal VOC 1-10 + 11-20}

\begin{table*}[!htb]
\centering
\caption{Results when learning the last 10 classes}\label{tab:results-min-voc-10}
\scalebox{0.5}{
\begin{tabular}{lrrrrrrrrrrrrrrrrrrrrrrrrr} 
\hline
\multicolumn{2}{c}{\textbf{\textcolor[rgb]{0,0.502,0}{10 + 10}}}                          & \multicolumn{1}{c}{aero} & \multicolumn{1}{c}{cycle} & \multicolumn{1}{c}{bird} & \multicolumn{1}{c}{boat} & \multicolumn{1}{c}{bottle} & \multicolumn{1}{c}{bus} & \multicolumn{1}{c}{car} & \multicolumn{1}{c}{cat} & \multicolumn{1}{c}{chair} & \multicolumn{1}{c}{cow} & \multicolumn{1}{c}{{\cellcolor[rgb]{0.812,0.886,0.953}}table} & \multicolumn{1}{c}{{\cellcolor[rgb]{0.812,0.886,0.953}}dog} & \multicolumn{1}{c}{{\cellcolor[rgb]{0.812,0.886,0.953}}horse} & \multicolumn{1}{c}{{\cellcolor[rgb]{0.812,0.886,0.953}}bike} & \multicolumn{1}{c}{{\cellcolor[rgb]{0.812,0.886,0.953}}person} & \multicolumn{1}{c}{{\cellcolor[rgb]{0.812,0.886,0.953}}plant} & \multicolumn{1}{c}{{\cellcolor[rgb]{0.812,0.886,0.953}}sheep} & \multicolumn{1}{c}{{\cellcolor[rgb]{0.812,0.886,0.953}}sofa} & \multicolumn{1}{c}{{\cellcolor[rgb]{0.812,0.886,0.953}}train} & \multicolumn{1}{c}{{\cellcolor[rgb]{0.812,0.886,0.953}}tv} & \multicolumn{1}{c}{\textbf{mAP}} & \multicolumn{1}{c}{$\Omega_{all} \uparrow$} & \multicolumn{1}{c}{\textbf{\textbf{RSD~}}($\%$) $\downarrow$} & \multicolumn{1}{c}{\textbf{\textbf{\textbf{\textbf{RPD~}}}}($\%$)~$\downarrow$}  \\ 
\hline
\multicolumn{2}{l}{Upper-bound}                                                           & 73.5                     & 80.6                      & 77.4                     & 61.2                     & 62.2                       & 79.9                    & 83.4                    & 86.7                    & 47.6                      & 78                      & {\cellcolor[rgb]{0.812,0.886,0.953}}68.1                      & {\cellcolor[rgb]{0.812,0.886,0.953}}85.1                    & {\cellcolor[rgb]{0.812,0.886,0.953}}83.7                      & {\cellcolor[rgb]{0.812,0.886,0.953}}82.8                     & {\cellcolor[rgb]{0.812,0.886,0.953}}79.1                       & {\cellcolor[rgb]{0.812,0.886,0.953}}42.5                      & {\cellcolor[rgb]{0.812,0.886,0.953}}75.7                      & {\cellcolor[rgb]{0.812,0.886,0.953}}64.9                     & {\cellcolor[rgb]{0.812,0.886,0.953}}79                        & {\cellcolor[rgb]{0.812,0.886,0.953}}76.2                   & 73.4                             & \multicolumn{1}{c}{-}                       & \multicolumn{1}{c}{-}                                         & \multicolumn{1}{c}{-}                                                            \\
\multicolumn{2}{l}{First 10}                                                              & 79.2                     & 85.6                      & 76.5                     & 66.7                     & 65.9                       & 78.9                    & 85.2                    & 86.6                    & 60.2                      & 84.7                    & {\cellcolor[rgb]{0.812,0.886,0.953}}0                         & {\cellcolor[rgb]{0.812,0.886,0.953}}0                       & {\cellcolor[rgb]{0.812,0.886,0.953}}0                         & {\cellcolor[rgb]{0.812,0.886,0.953}}0                        & {\cellcolor[rgb]{0.812,0.886,0.953}}0                          & {\cellcolor[rgb]{0.812,0.886,0.953}}0                         & {\cellcolor[rgb]{0.812,0.886,0.953}}0                         & {\cellcolor[rgb]{0.812,0.886,0.953}}0                        & {\cellcolor[rgb]{0.812,0.886,0.953}}0                         & {\cellcolor[rgb]{0.812,0.886,0.953}}0                      & 38.5                             & \multicolumn{1}{c}{-}                       & \multicolumn{1}{c}{-}                                         & \multicolumn{1}{c}{-}                                                            \\
\multicolumn{2}{l}{New 10}                                                                & 0                        & 0                         & 0                        & 0                        & 0                          & 0                       & 0                       & 0                       & 0                         & 0                       & {\cellcolor[rgb]{0.812,0.886,0.953}}74.6                      & {\cellcolor[rgb]{0.812,0.886,0.953}}85.7                    & {\cellcolor[rgb]{0.812,0.886,0.953}}86.1                      & {\cellcolor[rgb]{0.812,0.886,0.953}}79.9                     & {\cellcolor[rgb]{0.812,0.886,0.953}}79.8                       & {\cellcolor[rgb]{0.812,0.886,0.953}}43.9                      & {\cellcolor[rgb]{0.812,0.886,0.953}}76.3                      & {\cellcolor[rgb]{0.812,0.886,0.953}}68.5                     & {\cellcolor[rgb]{0.812,0.886,0.953}}80.5                      & {\cellcolor[rgb]{0.812,0.886,0.953}}76.3                   & 37.6                             & \multicolumn{1}{c}{-}                       & \multicolumn{1}{c}{-}                                         & \multicolumn{1}{c}{-}                                                            \\ 
\hline
\multicolumn{2}{l}{ILOD}                                                                  & 67.1                     & 64.1                      & 45.7                     & 40.9                     & 52.2                       & 66.5                    & 83.4                    & 75.3                    & 46.4                      & 59.4                    & {\cellcolor[rgb]{0.812,0.886,0.953}}64.1                      & {\cellcolor[rgb]{0.812,0.886,0.953}}74.8                    & {\cellcolor[rgb]{0.812,0.886,0.953}}77.1                      & {\cellcolor[rgb]{0.812,0.886,0.953}}67.1                     & {\cellcolor[rgb]{0.812,0.886,0.953}}63.3                       & {\cellcolor[rgb]{0.812,0.886,0.953}}32.7                      & {\cellcolor[rgb]{0.812,0.886,0.953}}61.3                      & {\cellcolor[rgb]{0.812,0.886,0.953}}56.8                     & {\cellcolor[rgb]{0.812,0.886,0.953}}73.7                      & {\cellcolor[rgb]{0.812,0.886,0.953}}67.3                   & 62.0                             & 0.84                                        & 17.65                                                         & 13.48                                                                            \\
\multicolumn{2}{l}{RILOD}                                                                 & 71.7                     & 81.7                      & 66.9                     & 49.6                     & 58                         & 65.9                    & 84.7                    & 76.8                    & 50.1                      & 69.4                    & {\cellcolor[rgb]{0.812,0.886,0.953}}67                        & {\cellcolor[rgb]{0.812,0.886,0.953}}72.8                    & {\cellcolor[rgb]{0.812,0.886,0.953}}77.3                      & {\cellcolor[rgb]{0.812,0.886,0.953}}73.8                     & {\cellcolor[rgb]{0.812,0.886,0.953}}74.9                       & {\cellcolor[rgb]{0.812,0.886,0.953}}39.9                      & {\cellcolor[rgb]{0.812,0.886,0.953}}68.5                      & {\cellcolor[rgb]{0.812,0.886,0.953}}61.5                     & {\cellcolor[rgb]{0.812,0.886,0.953}}75.5                      & {\cellcolor[rgb]{0.812,0.886,0.953}}72.4                   & 67.9                             & 0.93                                        & 7.59                                                          & 7.29                                                                             \\ 
\hline
\multirow{4}{*}{MMN}                                                                 & 25 & 59.2                     & 37.4                      & 38.7                     & 33.3                     & 17.2                       & 46.3                    & 52.9                    & 57.5                    & 5.9                       & 45.7                    & {\cellcolor[rgb]{0.812,0.886,0.953}}62.9                      & {\cellcolor[rgb]{0.812,0.886,0.953}}73.6                    & {\cellcolor[rgb]{0.812,0.886,0.953}}76                        & {\cellcolor[rgb]{0.812,0.886,0.953}}68.8                     & {\cellcolor[rgb]{0.812,0.886,0.953}}77.1                       & {\cellcolor[rgb]{0.812,0.886,0.953}}37.6                      & {\cellcolor[rgb]{0.812,0.886,0.953}}62.9                      & {\cellcolor[rgb]{0.812,0.886,0.953}}60.9                     & {\cellcolor[rgb]{0.812,0.886,0.953}}72.5                      & {\cellcolor[rgb]{0.812,0.886,0.953}}73.5                   & 53.0                             & 0.72                                        & 45.84                                                         & 9.72                                                                             \\
                                                                                     & 50 & 65.0                     & 42.7                      & 43.4                     & 37.6                     & 19.8                       & 53.1                    & 58.5                    & 58.5                    & 6.0                       & 46.0                    & {\cellcolor[rgb]{0.812,0.886,0.953}}59.4                      & {\cellcolor[rgb]{0.812,0.886,0.953}}72.6                    & {\cellcolor[rgb]{0.812,0.886,0.953}}73.1                      & {\cellcolor[rgb]{0.812,0.886,0.953}}69.5                     & {\cellcolor[rgb]{0.812,0.886,0.953}}75.5                       & {\cellcolor[rgb]{0.812,0.886,0.953}}35.7                      & {\cellcolor[rgb]{0.812,0.886,0.953}}60.0                      & {\cellcolor[rgb]{0.812,0.886,0.953}}59.2                     & {\cellcolor[rgb]{0.812,0.886,0.953}}69.2                      & {\cellcolor[rgb]{0.812,0.886,0.953}}71.7                   & 53.8                             & \textbf{0.73}                               & \textbf{40.89}                                                & \textbf{12.44}                                                                   \\
                                                                                     & 75 & 61.5                     & 40.3                      & 49.0                     & 35.8                     & 19.5                       & 48.0                    & 54.8                    & 52.3                    & 10.5                      & 44.0                    & {\cellcolor[rgb]{0.812,0.886,0.953}}62.5                      & {\cellcolor[rgb]{0.812,0.886,0.953}}71.0                    & {\cellcolor[rgb]{0.812,0.886,0.953}}74.1                      & {\cellcolor[rgb]{0.812,0.886,0.953}}68.4                     & {\cellcolor[rgb]{0.812,0.886,0.953}}75.6                       & {\cellcolor[rgb]{0.812,0.886,0.953}}36.2                      & {\cellcolor[rgb]{0.812,0.886,0.953}}59.6                      & {\cellcolor[rgb]{0.812,0.886,0.953}}61.3                     & {\cellcolor[rgb]{0.812,0.886,0.953}}69.6                      & {\cellcolor[rgb]{0.812,0.886,0.953}}70.7                   & 53.2                             & 0.73                                        & 42.91                                                         & 12.00                                                                            \\
                                                                                     & 90 & 67.2                     & 24.9                      & 56                       & 39.9                     & 31.2                       & 59.1                    & 62.2                    & 64.6                    & 6.5                       & 53.4                    & {\cellcolor[rgb]{0.812,0.886,0.953}}34.1                      & {\cellcolor[rgb]{0.812,0.886,0.953}}53.5                    & {\cellcolor[rgb]{0.812,0.886,0.953}}35.2                      & {\cellcolor[rgb]{0.812,0.886,0.953}}63.1                     & {\cellcolor[rgb]{0.812,0.886,0.953}}72.1                       & {\cellcolor[rgb]{0.812,0.886,0.953}}27.5                      & {\cellcolor[rgb]{0.812,0.886,0.953}}30                        & {\cellcolor[rgb]{0.812,0.886,0.953}}45.3                     & {\cellcolor[rgb]{0.812,0.886,0.953}}61.9                      & {\cellcolor[rgb]{0.812,0.886,0.953}}62.9                   & 47.5                             & 0.65                                        & 36.18                                                         & 34.27                                                                            \\ 
\hline\hline
\multirow{4}{*}{\begin{tabular}[c]{@{}l@{}}Gradient\\penalty \\of 1\%\end{tabular}}  & 25 & 59.2                     & 37.4                      & 38.5                     & 33.3                     & 17.1                       & 46.1                    & 52.8                    & 57.6                    & 5.9                       & 45.8                    & {\cellcolor[rgb]{0.812,0.886,0.953}}62.9                      & {\cellcolor[rgb]{0.812,0.886,0.953}}73.5                    & {\cellcolor[rgb]{0.812,0.886,0.953}}76.1                      & {\cellcolor[rgb]{0.812,0.886,0.953}}68.6                     & {\cellcolor[rgb]{0.812,0.886,0.953}}77.1                       & {\cellcolor[rgb]{0.812,0.886,0.953}}37.4                      & {\cellcolor[rgb]{0.812,0.886,0.953}}62.9                      & {\cellcolor[rgb]{0.812,0.886,0.953}}61                       & {\cellcolor[rgb]{0.812,0.886,0.953}}72.6                      & {\cellcolor[rgb]{0.812,0.886,0.953}}73.5                   & 53.0                             & 0.72                                        & 45.90                                                         & 9.74                                                                             \\
                                                                                     & 50 & 64.9                     & 43.9                      & 43.3                     & 37.2                     & 19.3                       & 53.1                    & 58.4                    & 58.4                    & 5.6                       & 46.0                    & {\cellcolor[rgb]{0.812,0.886,0.953}}59.3                      & {\cellcolor[rgb]{0.812,0.886,0.953}}72.7                    & {\cellcolor[rgb]{0.812,0.886,0.953}}73.1                      & {\cellcolor[rgb]{0.812,0.886,0.953}}69.6                     & {\cellcolor[rgb]{0.812,0.886,0.953}}75.6                       & {\cellcolor[rgb]{0.812,0.886,0.953}}35.8                      & {\cellcolor[rgb]{0.812,0.886,0.953}}60.2                      & {\cellcolor[rgb]{0.812,0.886,0.953}}59.2                     & {\cellcolor[rgb]{0.812,0.886,0.953}}69.4                      & {\cellcolor[rgb]{0.812,0.886,0.953}}71.8                   & 53.8                             & \textbf{0.73}                               & \textbf{40.91}                                                & \textbf{12.34}                                                                   \\
                                                                                     & 75 & 63.6                     & 41.0                      & 49.9                     & 36.7                     & 19.6                       & 48.4                    & 57.0                    & 53.0                    & 10.5                      & 43.9                    & {\cellcolor[rgb]{0.812,0.886,0.953}}61.9                      & {\cellcolor[rgb]{0.812,0.886,0.953}}71.5                    & {\cellcolor[rgb]{0.812,0.886,0.953}}74.3                      & {\cellcolor[rgb]{0.812,0.886,0.953}}67.9                     & {\cellcolor[rgb]{0.812,0.886,0.953}}75.4                       & {\cellcolor[rgb]{0.812,0.886,0.953}}35.8                      & {\cellcolor[rgb]{0.812,0.886,0.953}}59.5                      & {\cellcolor[rgb]{0.812,0.886,0.953}}61.1                     & {\cellcolor[rgb]{0.812,0.886,0.953}}69.4                      & {\cellcolor[rgb]{0.812,0.886,0.953}}70.4                   & 53.5                             & 0.73                                        & 41.84                                                         & 12.23                                                                            \\
                                                                                     & 90 & 67.2                     & 25.1                      & 55.2                     & 41                       & 30.1                       & 58.9                    & 62.2                    & 63.9                    & 5                         & 52.9                    & {\cellcolor[rgb]{0.812,0.886,0.953}}38.2                      & {\cellcolor[rgb]{0.812,0.886,0.953}}55                      & {\cellcolor[rgb]{0.812,0.886,0.953}}44.5                      & {\cellcolor[rgb]{0.812,0.886,0.953}}64.9                     & {\cellcolor[rgb]{0.812,0.886,0.953}}72.5                       & {\cellcolor[rgb]{0.812,0.886,0.953}}28.6                      & {\cellcolor[rgb]{0.812,0.886,0.953}}35                        & {\cellcolor[rgb]{0.812,0.886,0.953}}47.7                     & {\cellcolor[rgb]{0.812,0.886,0.953}}62.6                      & {\cellcolor[rgb]{0.812,0.886,0.953}}64.4                   & 48.7                             & 0.66                                        & 36.66                                                         & 30.49                                                                            \\ 
\hline
\multirow{4}{*}{\begin{tabular}[c]{@{}l@{}}Gradient\\penalty \\of 10\%\end{tabular}} & 25 & 59                       & 36.8                      & 36.5                     & 33                       & 16.5                       & 46                      & 52.7                    & 56.8                    & 5.8                       & 45.8                    & {\cellcolor[rgb]{0.812,0.886,0.953}}63.1                      & {\cellcolor[rgb]{0.812,0.886,0.953}}73.7                    & {\cellcolor[rgb]{0.812,0.886,0.953}}76.5                      & {\cellcolor[rgb]{0.812,0.886,0.953}}68.6                     & {\cellcolor[rgb]{0.812,0.886,0.953}}77.1                       & {\cellcolor[rgb]{0.812,0.886,0.953}}37.9                      & {\cellcolor[rgb]{0.812,0.886,0.953}}63.2                      & {\cellcolor[rgb]{0.812,0.886,0.953}}61.1                     & {\cellcolor[rgb]{0.812,0.886,0.953}}73                        & {\cellcolor[rgb]{0.812,0.886,0.953}}73.3                   & 52.8                             & 0.72                                        & 46.55                                                         & 9.48                                                                             \\
                                                                                     & 50 & 67.2                     & 44                        & 43.5                     & 38                       & 20.4                       & 51.8                    & 60.8                    & 60.5                    & 4.7                       & 46.5                    & {\cellcolor[rgb]{0.812,0.886,0.953}}59.1                      & {\cellcolor[rgb]{0.812,0.886,0.953}}72.7                    & {\cellcolor[rgb]{0.812,0.886,0.953}}73.2                      & {\cellcolor[rgb]{0.812,0.886,0.953}}68.9                     & {\cellcolor[rgb]{0.812,0.886,0.953}}75.6                       & {\cellcolor[rgb]{0.812,0.886,0.953}}34.7                      & {\cellcolor[rgb]{0.812,0.886,0.953}}59.6                      & {\cellcolor[rgb]{0.812,0.886,0.953}}59                       & {\cellcolor[rgb]{0.812,0.886,0.953}}69.8                      & {\cellcolor[rgb]{0.812,0.886,0.953}}71                     & 54.1                             & \textbf{0.74}                               & \textbf{39.94}                                                & \textbf{12.74}                                                                   \\
                                                                                     & 75 & 66.5                     & 44.1                      & 50.8                     & 37.0                     & 19.5                       & 52.1                    & 57.2                    & 56.1                    & 8.3                       & 46.2                    & {\cellcolor[rgb]{0.812,0.886,0.953}}60.4                      & {\cellcolor[rgb]{0.812,0.886,0.953}}70.2                    & {\cellcolor[rgb]{0.812,0.886,0.953}}73.0                      & {\cellcolor[rgb]{0.812,0.886,0.953}}68.7                     & {\cellcolor[rgb]{0.812,0.886,0.953}}75.4                       & {\cellcolor[rgb]{0.812,0.886,0.953}}35.4                      & {\cellcolor[rgb]{0.812,0.886,0.953}}59.3                      & {\cellcolor[rgb]{0.812,0.886,0.953}}58.7                     & {\cellcolor[rgb]{0.812,0.886,0.953}}69.3                      & {\cellcolor[rgb]{0.812,0.886,0.953}}70.9                   & 53.9                             & 0.73                                        & 39.93                                                         & 13.08                                                                            \\
                                                                                     & 90 & 67.6                     & 25.8                      & 50.6                     & 39.5                     & 24.9                       & 57.2                    & 61.5                    & 58.5                    & 4.7                       & 47.6                    & {\cellcolor[rgb]{0.812,0.886,0.953}}57.2                      & {\cellcolor[rgb]{0.812,0.886,0.953}}68.1                    & {\cellcolor[rgb]{0.812,0.886,0.953}}69.8                      & {\cellcolor[rgb]{0.812,0.886,0.953}}70.7                     & {\cellcolor[rgb]{0.812,0.886,0.953}}75.3                       & {\cellcolor[rgb]{0.812,0.886,0.953}}34.0                      & {\cellcolor[rgb]{0.812,0.886,0.953}}55.1                      & {\cellcolor[rgb]{0.812,0.886,0.953}}57.7                     & {\cellcolor[rgb]{0.812,0.886,0.953}}68.3                      & {\cellcolor[rgb]{0.812,0.886,0.953}}69.3                   & 53.2                             & 0.72                                        & 39.88                                                         & 15.24                                                                            \\ 
\hline\hline
\multirow{4}{*}{\begin{tabular}[c]{@{}l@{}}Freezing\\based on\\mean\end{tabular}}    & 25 & 63                       & 48.4                      & 57.3                     & 36.1                     & 19.9                       & 57.1                    & 49.8                    & 66                      & 7.7                       & 45                      & {\cellcolor[rgb]{0.812,0.886,0.953}}54                        & {\cellcolor[rgb]{0.812,0.886,0.953}}64                      & {\cellcolor[rgb]{0.812,0.886,0.953}}64                        & {\cellcolor[rgb]{0.812,0.886,0.953}}70.4                     & {\cellcolor[rgb]{0.812,0.886,0.953}}72.1                       & {\cellcolor[rgb]{0.812,0.886,0.953}}33.9                      & {\cellcolor[rgb]{0.812,0.886,0.953}}49.7                      & {\cellcolor[rgb]{0.812,0.886,0.953}}58.6                     & {\cellcolor[rgb]{0.812,0.886,0.953}}62.1                      & {\cellcolor[rgb]{0.812,0.886,0.953}}66.6                   & 52.3                             & 0.71                                        & 38.18                                                         & 19.31                                                                            \\
                                                                                     & 50 & 63.4                     & 48.6                      & 58                       & 39.1                     & 19                         & 57.4                    & 50                      & 66.2                    & 8.4                       & 44.3                    & {\cellcolor[rgb]{0.812,0.886,0.953}}53.8                      & {\cellcolor[rgb]{0.812,0.886,0.953}}63.3                    & {\cellcolor[rgb]{0.812,0.886,0.953}}63.8                      & {\cellcolor[rgb]{0.812,0.886,0.953}}70.3                     & {\cellcolor[rgb]{0.812,0.886,0.953}}72.2                       & {\cellcolor[rgb]{0.812,0.886,0.953}}33.2                      & {\cellcolor[rgb]{0.812,0.886,0.953}}49.8                      & {\cellcolor[rgb]{0.812,0.886,0.953}}58.5                     & {\cellcolor[rgb]{0.812,0.886,0.953}}61.6                      & {\cellcolor[rgb]{0.812,0.886,0.953}}67.1                   & 52.4                             & \textbf{0.71}                               & \textbf{37.63}                                                & \textbf{19.56}                                                                   \\
                                                                                     & 75 & 58.8                     & 49.1                      & 55.6                     & 41.1                     & 17.5                       & 58.1                    & 43.5                    & 67.5                    & 11                        & 43.3                    & {\cellcolor[rgb]{0.812,0.886,0.953}}47                        & {\cellcolor[rgb]{0.812,0.886,0.953}}66                      & {\cellcolor[rgb]{0.812,0.886,0.953}}54.3                      & {\cellcolor[rgb]{0.812,0.886,0.953}}70                       & {\cellcolor[rgb]{0.812,0.886,0.953}}70.2                       & {\cellcolor[rgb]{0.812,0.886,0.953}}32.4                      & {\cellcolor[rgb]{0.812,0.886,0.953}}47.4                      & {\cellcolor[rgb]{0.812,0.886,0.953}}58.8                     & {\cellcolor[rgb]{0.812,0.886,0.953}}51                        & {\cellcolor[rgb]{0.812,0.886,0.953}}67.5                   & 50.5                             & 0.69                                        & 38.84                                                         & 23.51                                                                            \\
                                                                                     & 90 & 54.2                     & 49.7                      & 51.2                     & 39.8                     & 23.9                       & 60.1                    & 44.1                    & 70.7                    & 14.2                      & 46.6                    & {\cellcolor[rgb]{0.812,0.886,0.953}}24.1                      & {\cellcolor[rgb]{0.812,0.886,0.953}}57.9                    & {\cellcolor[rgb]{0.812,0.886,0.953}}46.7                      & {\cellcolor[rgb]{0.812,0.886,0.953}}63.5                     & {\cellcolor[rgb]{0.812,0.886,0.953}}59.3                       & {\cellcolor[rgb]{0.812,0.886,0.953}}28.8                      & {\cellcolor[rgb]{0.812,0.886,0.953}}42                        & {\cellcolor[rgb]{0.812,0.886,0.953}}58.4                     & {\cellcolor[rgb]{0.812,0.886,0.953}}43.8                      & {\cellcolor[rgb]{0.812,0.886,0.953}}59.4                   & 46.9                             & 0.64                                        & 37.61                                                         & 34.51                                                                            \\ 
\hline
\multirow{4}{*}{\begin{tabular}[c]{@{}l@{}}Freezing\\based on\\median\end{tabular}}  & 25 & 60.9                     & 48.3                      & 57.8                     & 34.3                     & 23                         & 57.3                    & 43.8                    & 65.7                    & 10.4                      & 46.2                    & {\cellcolor[rgb]{0.812,0.886,0.953}}55.1                      & {\cellcolor[rgb]{0.812,0.886,0.953}}65.2                    & {\cellcolor[rgb]{0.812,0.886,0.953}}67.7                      & {\cellcolor[rgb]{0.812,0.886,0.953}}71.3                     & {\cellcolor[rgb]{0.812,0.886,0.953}}72.8                       & {\cellcolor[rgb]{0.812,0.886,0.953}}33.9                      & {\cellcolor[rgb]{0.812,0.886,0.953}}52.8                      & {\cellcolor[rgb]{0.812,0.886,0.953}}59.3                     & {\cellcolor[rgb]{0.812,0.886,0.953}}65                        & {\cellcolor[rgb]{0.812,0.886,0.953}}68.3                   & 53.0                             & \textbf{0.72}                               & \textbf{38.54}                                                & \textbf{17.13}                                                                   \\
                                                                                     & 50 & 58.5                     & 48.8                      & 55.4                     & 41.5                     & 18.7                       & 58.4                    & 43.8                    & 70.5                    & 11                        & 41.9                    & {\cellcolor[rgb]{0.812,0.886,0.953}}53.7                      & {\cellcolor[rgb]{0.812,0.886,0.953}}66.8                    & {\cellcolor[rgb]{0.812,0.886,0.953}}54.2                      & {\cellcolor[rgb]{0.812,0.886,0.953}}71.2                     & {\cellcolor[rgb]{0.812,0.886,0.953}}71.8                       & {\cellcolor[rgb]{0.812,0.886,0.953}}35.1                      & {\cellcolor[rgb]{0.812,0.886,0.953}}49.4                      & {\cellcolor[rgb]{0.812,0.886,0.953}}59.6                     & {\cellcolor[rgb]{0.812,0.886,0.953}}52.6                      & {\cellcolor[rgb]{0.812,0.886,0.953}}68.7                   & 51.6                             & 0.70                                        & 38.43                                                         & 20.99                                                                            \\
                                                                                     & 75 & 54.6                     & 48.9                      & 52.7                     & 38.4                     & 24.6                       & 59.3                    & 44.1                    & 70.9                    & 14.1                      & 47.2                    & {\cellcolor[rgb]{0.812,0.886,0.953}}29.4                      & {\cellcolor[rgb]{0.812,0.886,0.953}}58.7                    & {\cellcolor[rgb]{0.812,0.886,0.953}}49.5                      & {\cellcolor[rgb]{0.812,0.886,0.953}}63.6                     & {\cellcolor[rgb]{0.812,0.886,0.953}}60.4                       & {\cellcolor[rgb]{0.812,0.886,0.953}}29                        & {\cellcolor[rgb]{0.812,0.886,0.953}}42.8                      & {\cellcolor[rgb]{0.812,0.886,0.953}}58.6                     & {\cellcolor[rgb]{0.812,0.886,0.953}}45.8                      & {\cellcolor[rgb]{0.812,0.886,0.953}}59.9                   & 47.6                             & 0.65                                        & 37.57                                                         & 32.62                                                                            \\
                                                                                     & 90 & 53.6                     & 42.4                      & 51.9                     & 38                       & 23.8                       & 60.1                    & 44.1                    & 71.3                    & 14.4                      & 47.5                    & {\cellcolor[rgb]{0.812,0.886,0.953}}28                        & {\cellcolor[rgb]{0.812,0.886,0.953}}58.7                    & {\cellcolor[rgb]{0.812,0.886,0.953}}49                        & {\cellcolor[rgb]{0.812,0.886,0.953}}64.7                     & {\cellcolor[rgb]{0.812,0.886,0.953}}60.1                       & {\cellcolor[rgb]{0.812,0.886,0.953}}25.4                      & {\cellcolor[rgb]{0.812,0.886,0.953}}42.3                      & {\cellcolor[rgb]{0.812,0.886,0.953}}58.4                     & {\cellcolor[rgb]{0.812,0.886,0.953}}46.8                      & {\cellcolor[rgb]{0.812,0.886,0.953}}59.7                   & 47.0                             & 0.64                                        & 38.62                                                         & 33.25                                                                            \\ 
\hline
\multirow{4}{*}{\begin{tabular}[c]{@{}l@{}}Freezing\\based on\\std\end{tabular}}     & 25 & 62.7                     & 48.5                      & 57.4                     & 36.2                     & 19.6                       & 57.1                    & 49.8                    & 66.1                    & 7.6                       & 45.2                    & {\cellcolor[rgb]{0.812,0.886,0.953}}54.1                      & {\cellcolor[rgb]{0.812,0.886,0.953}}64.1                    & {\cellcolor[rgb]{0.812,0.886,0.953}}64                        & {\cellcolor[rgb]{0.812,0.886,0.953}}70.2                     & {\cellcolor[rgb]{0.812,0.886,0.953}}72.2                       & {\cellcolor[rgb]{0.812,0.886,0.953}}33.9                      & {\cellcolor[rgb]{0.812,0.886,0.953}}49.8                      & {\cellcolor[rgb]{0.812,0.886,0.953}}58.4                     & {\cellcolor[rgb]{0.812,0.886,0.953}}62.1                      & {\cellcolor[rgb]{0.812,0.886,0.953}}66.4                   & 52.3                             & \textbf{0.71}                               & \textbf{38.20}                                                & \textbf{19.34}                                                                   \\
                                                                                     & 50 & 62.6                     & 48.4                      & 56.8                     & 38.5                     & 19.2                       & 57.8                    & 50                      & 65.9                    & 7                         & 45.1                    & {\cellcolor[rgb]{0.812,0.886,0.953}}52.9                      & {\cellcolor[rgb]{0.812,0.886,0.953}}63.8                    & {\cellcolor[rgb]{0.812,0.886,0.953}}63.7                      & {\cellcolor[rgb]{0.812,0.886,0.953}}70.2                     & {\cellcolor[rgb]{0.812,0.886,0.953}}71.8                       & {\cellcolor[rgb]{0.812,0.886,0.953}}32.8                      & {\cellcolor[rgb]{0.812,0.886,0.953}}49.9                      & {\cellcolor[rgb]{0.812,0.886,0.953}}57.7                     & {\cellcolor[rgb]{0.812,0.886,0.953}}60.7                      & {\cellcolor[rgb]{0.812,0.886,0.953}}66.4                   & 52.1                             & 0.71                                        & 38.05                                                         & 20.06                                                                            \\
                                                                                     & 75 & 62.1                     & 47.3                      & 57.8                     & 38.8                     & 19.5                       & 58.2                    & 50.1                    & 65.3                    & 8.5                       & 44.6                    & {\cellcolor[rgb]{0.812,0.886,0.953}}53.4                      & {\cellcolor[rgb]{0.812,0.886,0.953}}62.7                    & {\cellcolor[rgb]{0.812,0.886,0.953}}64                        & {\cellcolor[rgb]{0.812,0.886,0.953}}69.9                     & {\cellcolor[rgb]{0.812,0.886,0.953}}71.5                       & {\cellcolor[rgb]{0.812,0.886,0.953}}31.7                      & {\cellcolor[rgb]{0.812,0.886,0.953}}51.1                      & {\cellcolor[rgb]{0.812,0.886,0.953}}57.1                     & {\cellcolor[rgb]{0.812,0.886,0.953}}60.8                      & {\cellcolor[rgb]{0.812,0.886,0.953}}65.1                   & 52.0                             & 0.71                                        & 37.93                                                         & 20.41                                                                            \\
                                                                                     & 90 & 57.2                     & 40.8                      & 55                       & 29.8                     & 11.5                       & 57.3                    & 44.2                    & 65.5                    & 10.8                      & 41.7                    & {\cellcolor[rgb]{0.812,0.886,0.953}}39.6                      & {\cellcolor[rgb]{0.812,0.886,0.953}}58.9                    & {\cellcolor[rgb]{0.812,0.886,0.953}}55.3                      & {\cellcolor[rgb]{0.812,0.886,0.953}}62.2                     & {\cellcolor[rgb]{0.812,0.886,0.953}}68.9                       & {\cellcolor[rgb]{0.812,0.886,0.953}}33.3                      & {\cellcolor[rgb]{0.812,0.886,0.953}}55.2                      & {\cellcolor[rgb]{0.812,0.886,0.953}}60                       & {\cellcolor[rgb]{0.812,0.886,0.953}}54.4                      & {\cellcolor[rgb]{0.812,0.886,0.953}}64.1                   & 48.3                             & 0.66                                        & 43.16                                                         & 25.24                                                                            \\ 
\hline
\multirow{4}{*}{\begin{tabular}[c]{@{}l@{}}Freezing\\based on\\entropy\end{tabular}} & 25 & 68.3                     & 42.3                      & 49.8                     & 42.1                     & 15.3                       & 53.3                    & 60.8                    & 60.9                    & 4.8                       & 51.4                    & {\cellcolor[rgb]{0.812,0.886,0.953}}49.9                      & {\cellcolor[rgb]{0.812,0.886,0.953}}71.4                    & {\cellcolor[rgb]{0.812,0.886,0.953}}72.4                      & {\cellcolor[rgb]{0.812,0.886,0.953}}71                       & {\cellcolor[rgb]{0.812,0.886,0.953}}75.5                       & {\cellcolor[rgb]{0.812,0.886,0.953}}36.2                      & {\cellcolor[rgb]{0.812,0.886,0.953}}53.5                      & {\cellcolor[rgb]{0.812,0.886,0.953}}57.5                     & {\cellcolor[rgb]{0.812,0.886,0.953}}70.4                      & {\cellcolor[rgb]{0.812,0.886,0.953}}70.2                   & 53.9                             & \textbf{0.73}                               & \textbf{38.36}                                                & \textbf{14.87}                                                                   \\
                                                                                     & 50 & 60.8                     & 34.1                      & 48.2                     & 30.1                     & 32                         & 51.8                    & 42.2                    & 56.9                    & 14.9                      & 45.3                    & {\cellcolor[rgb]{0.812,0.886,0.953}}55.7                      & {\cellcolor[rgb]{0.812,0.886,0.953}}63                      & {\cellcolor[rgb]{0.812,0.886,0.953}}67.5                      & {\cellcolor[rgb]{0.812,0.886,0.953}}66.5                     & {\cellcolor[rgb]{0.812,0.886,0.953}}73                         & {\cellcolor[rgb]{0.812,0.886,0.953}}32.5                      & {\cellcolor[rgb]{0.812,0.886,0.953}}46.9                      & {\cellcolor[rgb]{0.812,0.886,0.953}}58.8                     & {\cellcolor[rgb]{0.812,0.886,0.953}}62.3                      & {\cellcolor[rgb]{0.812,0.886,0.953}}67.4                   & 50.5                             & 0.69                                        & 42.82                                                         & 19.56                                                                            \\
                                                                                     & 75 & 61.2                     & 31.9                      & 49.4                     & 32.8                     & 29.2                       & 55.7                    & 46.5                    & 57.4                    & 10.6                      & 47.7                    & {\cellcolor[rgb]{0.812,0.886,0.953}}55.8                      & {\cellcolor[rgb]{0.812,0.886,0.953}}66.6                    & {\cellcolor[rgb]{0.812,0.886,0.953}}65.4                      & {\cellcolor[rgb]{0.812,0.886,0.953}}64.5                     & {\cellcolor[rgb]{0.812,0.886,0.953}}71.8                       & {\cellcolor[rgb]{0.812,0.886,0.953}}30.8                      & {\cellcolor[rgb]{0.812,0.886,0.953}}45.7                      & {\cellcolor[rgb]{0.812,0.886,0.953}}57.7                     & {\cellcolor[rgb]{0.812,0.886,0.953}}63.8                      & {\cellcolor[rgb]{0.812,0.886,0.953}}66.4                   & 50.5                             & 0.69                                        & 41.99                                                         & 20.25                                                                            \\
                                                                                     & 90 & 54.6                     & 53.6                      & 63.8                     & 46.0                     & 24.4                       & 55.9                    & 53.4                    & 69.4                    & 20.0                      & 51.6                    & {\cellcolor[rgb]{0.812,0.886,0.953}}31.4                      & {\cellcolor[rgb]{0.812,0.886,0.953}}53.7                    & {\cellcolor[rgb]{0.812,0.886,0.953}}49.1                      & {\cellcolor[rgb]{0.812,0.886,0.953}}59.2                     & {\cellcolor[rgb]{0.812,0.886,0.953}}40.0                       & {\cellcolor[rgb]{0.812,0.886,0.953}}7.5                       & {\cellcolor[rgb]{0.812,0.886,0.953}}31.0                      & {\cellcolor[rgb]{0.812,0.886,0.953}}55.0                     & {\cellcolor[rgb]{0.812,0.886,0.953}}41.1                      & {\cellcolor[rgb]{0.812,0.886,0.953}}34.8                   & 44.8                             & 0.61                                        & 32.43                                                         & 45.58                                                                            \\
\hline
\end{tabular}}
\end{table*}

Table \ref{tab:results-min-voc-10} reports the results for the $10+10$ alternative. For this scenario, the final $mAP$ and $\Omega_{all}$ became more relevant as there was an equal representation of classes for their calculations. Results for applying a penalty to the gradient of selected neurons showed a slightly superior performance compared to completely freezing them. This was especially true in all scenarios where a 10\% penalty was applied. For this benchmark, freezing 25\% of the layers based on information entropy yielded the best results, followed by using the median of the activations to the same percentage of frozen layers. However, the final $mAP$ and $\Omega_{all}$ indicate that these simply arranged strategies might have a difficult time competing against traditional methods when processing a benchmark with more complexities. Nonetheless, they can still serve as a quick and strong baseline when compared to fine-tuning and MMN due to ease of implementation.

Overall for the $10+10$ scenario, all evaluated strategies produced comparable final in terms of $mAP$ and $\Omega_{all}$. Nevertheless, the best outcomes were observed when freezing or penalizing 50\% or less of the parameters. Since most detectors based on deep neural networks are overparameterized and not optimized directly for sparse connections, freezing more than 50\% of available parameters or layers might affect highly the network capacity for learning new objects. We believe this to be true mainly for learning new tasks with imbalanced category sets and objects that do not present visual similarities with the ones previously learned. The Incremental Pascal VOC benchmark presents not only an imbalanced occurrence of each category but also a considerable semantic difference for the labels of the two tasks, with the first having more instances from outdoor environments and the second focusing on instances from indoor scenes. This can be further investigated by exploring task-relatedness as a way to define the parameters that determine how layer-freezing should take place between updates.

Interestingly, as also shown in the final evaluation remarks of the PackNet strategy for classification, the final performance of the incremental model can be weakened since it only uses a fraction of the entire parameter set to learn new tasks \citep{delange2021continual}. However, this tradeoff is necessary to ensure stable performance in the tasks that were initially learned. Considering the necessity for quick adaptation in constrained environments, having a hyperparameter to adjust the plasticity of the model can be used as a feature to preserve the performance in previous scenarios and slightly adjust the network to the new circumstances. This feature can be especially beneficial when new updates with mixed data (i.e., old and new samples) are expected in the future.

\subsection{TAESA Benchmark}

\begin{table*}[!htb]
\centering
\caption{Results for incremental training on the TAESA Benchmark}\label{tab:taesa-results}
\scalebox{0.55}{
\begin{tabular}{c|crrrrrrrrrrrrrrr} 
\cline{4-17}
\multicolumn{3}{l}{}                                                    & \multicolumn{2}{c}{Task 1}                                                              & \multicolumn{2}{c}{Task 2}                                                              & \multicolumn{2}{c}{Task 3}                                                              & \multicolumn{2}{c}{Task 4}                                                              & \multicolumn{6}{c}{\textbf{Final Eval}}                                                                                                                                                                                                                                                                                                                                                                                                                                                                                                                         \\ 
\hline
\multicolumn{1}{l}{}                  & \%                    & Feature & {\cellcolor[rgb]{0.851,0.918,0.827}}mAP  & {\cellcolor[rgb]{0.812,0.886,0.953}}mAP[.50] & {\cellcolor[rgb]{0.851,0.918,0.827}}mAP  & {\cellcolor[rgb]{0.812,0.886,0.953}}mAP[.50] & {\cellcolor[rgb]{0.851,0.918,0.827}}mAP  & {\cellcolor[rgb]{0.812,0.886,0.953}}mAP[.50] & {\cellcolor[rgb]{0.851,0.918,0.827}}mAP  & {\cellcolor[rgb]{0.812,0.886,0.953}}mAP[.50] & {\cellcolor[rgb]{0.851,0.918,0.827}}\begin{tabular}[c]{@{}>{\cellcolor[rgb]{0.851,0.918,0.827}}r@{}}\textbf{Average }\\\textbf{mAP}\end{tabular} & {\cellcolor[rgb]{0.812,0.886,0.953}}\begin{tabular}[c]{@{}>{\cellcolor[rgb]{0.812,0.886,0.953}}r@{}}\textbf{Average }\\\textbf{mAP [.50]}\end{tabular} & {\cellcolor[rgb]{0.851,0.918,0.827}}$\Omega_mAP \uparrow$ & {\cellcolor[rgb]{0.812,0.886,0.953}}$\Omega_mAP[.50] \uparrow$ & {\cellcolor[rgb]{0.851,0.918,0.827}}$RSD_{mAP}\downarrow$ & {\cellcolor[rgb]{0.851,0.918,0.827}}$RPD_{mAP}\downarrow$  \\ 
\hline
\multirow{16}{*}{Freeze}              & \multirow{4}{*}{25}   & mean    & {\cellcolor[rgb]{0.851,0.918,0.827}}43.7 & {\cellcolor[rgb]{0.812,0.886,0.953}}67.9     & {\cellcolor[rgb]{0.851,0.918,0.827}}5.6  & {\cellcolor[rgb]{0.812,0.886,0.953}}13.5     & {\cellcolor[rgb]{0.851,0.918,0.827}}13.3 & {\cellcolor[rgb]{0.812,0.886,0.953}}24.1     & {\cellcolor[rgb]{0.851,0.918,0.827}}35.1 & {\cellcolor[rgb]{0.812,0.886,0.953}}60.8     & {\cellcolor[rgb]{0.851,0.918,0.827}}24.4                                                                                                         & {\cellcolor[rgb]{0.812,0.886,0.953}}41.6                                                                                                               & {\cellcolor[rgb]{0.851,0.918,0.827}}0.55                  & {\cellcolor[rgb]{0.812,0.886,0.953}}0.60                       & {\cellcolor[rgb]{0.851,0.918,0.827}}51.18                 & {\cellcolor[rgb]{0.851,0.918,0.827}}28.22                  \\
                                      &                       & median  & {\cellcolor[rgb]{0.851,0.918,0.827}}43.8 & {\cellcolor[rgb]{0.812,0.886,0.953}}65.4     & {\cellcolor[rgb]{0.851,0.918,0.827}}9.7  & {\cellcolor[rgb]{0.812,0.886,0.953}}21       & {\cellcolor[rgb]{0.851,0.918,0.827}}15.2 & {\cellcolor[rgb]{0.812,0.886,0.953}}36.9     & {\cellcolor[rgb]{0.851,0.918,0.827}}37.9 & {\cellcolor[rgb]{0.812,0.886,0.953}}64.5     & {\cellcolor[rgb]{0.851,0.918,0.827}}26.6                                                                                                         & {\cellcolor[rgb]{0.812,0.886,0.953}}47.0                                                                                                               & {\cellcolor[rgb]{0.851,0.918,0.827}}0.60                  & {\cellcolor[rgb]{0.812,0.886,0.953}}0.67                       & {\cellcolor[rgb]{0.851,0.918,0.827}}46.48                 & {\cellcolor[rgb]{0.851,0.918,0.827}}22.49                  \\
                                      &                       & std     & {\cellcolor[rgb]{0.851,0.918,0.827}}41.7 & {\cellcolor[rgb]{0.812,0.886,0.953}}62.5     & {\cellcolor[rgb]{0.851,0.918,0.827}}10.5 & {\cellcolor[rgb]{0.812,0.886,0.953}}21.6     & {\cellcolor[rgb]{0.851,0.918,0.827}}19.3 & {\cellcolor[rgb]{0.812,0.886,0.953}}32.9     & {\cellcolor[rgb]{0.851,0.918,0.827}}38.6 & {\cellcolor[rgb]{0.812,0.886,0.953}}64.9     & {\cellcolor[rgb]{0.851,0.918,0.827}}27.5                                                                                                         & {\cellcolor[rgb]{0.812,0.886,0.953}}45.5                                                                                                               & {\cellcolor[rgb]{0.851,0.918,0.827}}0.62                  & {\cellcolor[rgb]{0.812,0.886,0.953}}0.65                       & {\cellcolor[rgb]{0.851,0.918,0.827}}44.28                 & {\cellcolor[rgb]{0.851,0.918,0.827}}21.06                  \\
                                      &                       & entropy & {\cellcolor[rgb]{0.851,0.918,0.827}}41.2 & {\cellcolor[rgb]{0.812,0.886,0.953}}61.4     & {\cellcolor[rgb]{0.851,0.918,0.827}}15.6 & {\cellcolor[rgb]{0.812,0.886,0.953}}30.3     & {\cellcolor[rgb]{0.851,0.918,0.827}}21   & {\cellcolor[rgb]{0.812,0.886,0.953}}34.7     & {\cellcolor[rgb]{0.851,0.918,0.827}}39.8 & {\cellcolor[rgb]{0.812,0.886,0.953}}67.1     & {\cellcolor[rgb]{0.851,0.918,0.827}}\textbf{29.4}                                                                                                & {\cellcolor[rgb]{0.812,0.886,0.953}}\textbf{48.4}                                                                                                      & {\cellcolor[rgb]{0.851,0.918,0.827}}\textbf{0.66}         & {\cellcolor[rgb]{0.812,0.886,0.953}}\textbf{0.69}              & {\cellcolor[rgb]{0.851,0.918,0.827}}\textbf{39.33}        & {\cellcolor[rgb]{0.851,0.918,0.827}}\textbf{18.61}         \\ 
\hhline{~----------------}
                                      & \multirow{4}{*}{50}   & mean    & {\cellcolor[rgb]{0.851,0.918,0.827}}44.0 & {\cellcolor[rgb]{0.812,0.886,0.953}}69.6     & {\cellcolor[rgb]{0.851,0.918,0.827}}5.8  & {\cellcolor[rgb]{0.812,0.886,0.953}}13.9     & {\cellcolor[rgb]{0.851,0.918,0.827}}11.8 & {\cellcolor[rgb]{0.812,0.886,0.953}}23.2     & {\cellcolor[rgb]{0.851,0.918,0.827}}35   & {\cellcolor[rgb]{0.812,0.886,0.953}}61       & {\cellcolor[rgb]{0.851,0.918,0.827}}24.2                                                                                                         & {\cellcolor[rgb]{0.812,0.886,0.953}}41.9                                                                                                               & {\cellcolor[rgb]{0.851,0.918,0.827}}0.55                  & {\cellcolor[rgb]{0.812,0.886,0.953}}0.60                       & {\cellcolor[rgb]{0.851,0.918,0.827}}51.96                 & {\cellcolor[rgb]{0.851,0.918,0.827}}28.43                  \\
                                      &                       & median  & {\cellcolor[rgb]{0.851,0.918,0.827}}43.3 & {\cellcolor[rgb]{0.812,0.886,0.953}}64.7     & {\cellcolor[rgb]{0.851,0.918,0.827}}10.5 & {\cellcolor[rgb]{0.812,0.886,0.953}}22.5     & {\cellcolor[rgb]{0.851,0.918,0.827}}14.8 & {\cellcolor[rgb]{0.812,0.886,0.953}}26.3     & {\cellcolor[rgb]{0.851,0.918,0.827}}37.2 & {\cellcolor[rgb]{0.812,0.886,0.953}}62.6     & {\cellcolor[rgb]{0.851,0.918,0.827}}26.5                                                                                                         & {\cellcolor[rgb]{0.812,0.886,0.953}}44.0                                                                                                               & {\cellcolor[rgb]{0.851,0.918,0.827}}0.60                  & {\cellcolor[rgb]{0.812,0.886,0.953}}0.63                       & {\cellcolor[rgb]{0.851,0.918,0.827}}46.52                 & {\cellcolor[rgb]{0.851,0.918,0.827}}23.93                  \\
                                      &                       & std     & {\cellcolor[rgb]{0.851,0.918,0.827}}41.4 & {\cellcolor[rgb]{0.812,0.886,0.953}}64.4     & {\cellcolor[rgb]{0.851,0.918,0.827}}10.9 & {\cellcolor[rgb]{0.812,0.886,0.953}}22.8     & {\cellcolor[rgb]{0.851,0.918,0.827}}19.8 & {\cellcolor[rgb]{0.812,0.886,0.953}}34.3     & {\cellcolor[rgb]{0.851,0.918,0.827}}38.4 & {\cellcolor[rgb]{0.812,0.886,0.953}}64.9     & {\cellcolor[rgb]{0.851,0.918,0.827}}27.6                                                                                                         & {\cellcolor[rgb]{0.812,0.886,0.953}}46.6                                                                                                               & {\cellcolor[rgb]{0.851,0.918,0.827}}0.62                  & {\cellcolor[rgb]{0.812,0.886,0.953}}0.67                       & {\cellcolor[rgb]{0.851,0.918,0.827}}43.77                 & {\cellcolor[rgb]{0.851,0.918,0.827}}21.47                  \\
                                      &                       & entropy & {\cellcolor[rgb]{0.851,0.918,0.827}}41.0 & {\cellcolor[rgb]{0.812,0.886,0.953}}61.8     & {\cellcolor[rgb]{0.851,0.918,0.827}}16.6 & {\cellcolor[rgb]{0.812,0.886,0.953}}31.5     & {\cellcolor[rgb]{0.851,0.918,0.827}}22.2 & {\cellcolor[rgb]{0.812,0.886,0.953}}37.8     & {\cellcolor[rgb]{0.851,0.918,0.827}}39   & {\cellcolor[rgb]{0.812,0.886,0.953}}65.9     & {\cellcolor[rgb]{0.851,0.918,0.827}}\textbf{29.7}                                                                                                & {\cellcolor[rgb]{0.812,0.886,0.953}}\textbf{49.2}                                                                                                      & {\cellcolor[rgb]{0.851,0.918,0.827}}\textbf{0.67}         & {\cellcolor[rgb]{0.812,0.886,0.953}}\textbf{0.71}              & {\cellcolor[rgb]{0.851,0.918,0.827}}\textbf{37.77}        & {\cellcolor[rgb]{0.851,0.918,0.827}}\textbf{20.25}         \\ 
\hhline{~----------------}
                                      & \multirow{4}{*}{75}   & mean    & {\cellcolor[rgb]{0.851,0.918,0.827}}47.9 & {\cellcolor[rgb]{0.812,0.886,0.953}}71.4     & {\cellcolor[rgb]{0.851,0.918,0.827}}3.5  & {\cellcolor[rgb]{0.812,0.886,0.953}}9.8      & {\cellcolor[rgb]{0.851,0.918,0.827}}12.4 & {\cellcolor[rgb]{0.812,0.886,0.953}}24.1     & {\cellcolor[rgb]{0.851,0.918,0.827}}31   & {\cellcolor[rgb]{0.812,0.886,0.953}}55.3     & {\cellcolor[rgb]{0.851,0.918,0.827}}\textbf{31.4}                                                                                                & {\cellcolor[rgb]{0.812,0.886,0.953}}\textbf{49.0}                                                                                                      & {\cellcolor[rgb]{0.851,0.918,0.827}}\textbf{0.71}         & {\cellcolor[rgb]{0.812,0.886,0.953}}\textbf{0.70}              & {\cellcolor[rgb]{0.851,0.918,0.827}}50.28                 & {\cellcolor[rgb]{0.851,0.918,0.827}}36.61                  \\
                                      &                       & median  & {\cellcolor[rgb]{0.851,0.918,0.827}}45.9 & {\cellcolor[rgb]{0.812,0.886,0.953}}65.3     & {\cellcolor[rgb]{0.851,0.918,0.827}}6.8  & {\cellcolor[rgb]{0.812,0.886,0.953}}17.5     & {\cellcolor[rgb]{0.851,0.918,0.827}}17.4 & {\cellcolor[rgb]{0.812,0.886,0.953}}30.6     & {\cellcolor[rgb]{0.851,0.918,0.827}}32.9 & {\cellcolor[rgb]{0.812,0.886,0.953}}60       & {\cellcolor[rgb]{0.851,0.918,0.827}}30.9                                                                                                         & {\cellcolor[rgb]{0.812,0.886,0.953}}48.7                                                                                                               & {\cellcolor[rgb]{0.851,0.918,0.827}}0.70                  & {\cellcolor[rgb]{0.812,0.886,0.953}}0.70                       & {\cellcolor[rgb]{0.851,0.918,0.827}}45.37                 & {\cellcolor[rgb]{0.851,0.918,0.827}}32.72                  \\
                                      &                       & std     & {\cellcolor[rgb]{0.851,0.918,0.827}}44.1 & {\cellcolor[rgb]{0.812,0.886,0.953}}63.2     & {\cellcolor[rgb]{0.851,0.918,0.827}}10.8 & {\cellcolor[rgb]{0.812,0.886,0.953}}24       & {\cellcolor[rgb]{0.851,0.918,0.827}}19.3 & {\cellcolor[rgb]{0.812,0.886,0.953}}32.5     & {\cellcolor[rgb]{0.851,0.918,0.827}}34.4 & {\cellcolor[rgb]{0.812,0.886,0.953}}62.1     & {\cellcolor[rgb]{0.851,0.918,0.827}}30.5                                                                                                         & {\cellcolor[rgb]{0.812,0.886,0.953}}48.7                                                                                                               & {\cellcolor[rgb]{0.851,0.918,0.827}}0.69                  & {\cellcolor[rgb]{0.812,0.886,0.953}}0.70                       & {\cellcolor[rgb]{0.851,0.918,0.827}}42.14                 & {\cellcolor[rgb]{0.851,0.918,0.827}}29.65                  \\
                                      &                       & entropy & {\cellcolor[rgb]{0.851,0.918,0.827}}43.7 & {\cellcolor[rgb]{0.812,0.886,0.953}}63.1     & {\cellcolor[rgb]{0.851,0.918,0.827}}11.6 & {\cellcolor[rgb]{0.812,0.886,0.953}}21.9     & {\cellcolor[rgb]{0.851,0.918,0.827}}22.5 & {\cellcolor[rgb]{0.812,0.886,0.953}}38.5     & {\cellcolor[rgb]{0.851,0.918,0.827}}36.6 & {\cellcolor[rgb]{0.812,0.886,0.953}}62.3     & {\cellcolor[rgb]{0.851,0.918,0.827}}30.4                                                                                                         & {\cellcolor[rgb]{0.812,0.886,0.953}}48.7                                                                                                               & {\cellcolor[rgb]{0.851,0.918,0.827}}0.69                  & {\cellcolor[rgb]{0.812,0.886,0.953}}0.70                       & {\cellcolor[rgb]{0.851,0.918,0.827}}\textbf{39.33}        & {\cellcolor[rgb]{0.851,0.918,0.827}}\textbf{25.15}         \\ 
\hhline{~----------------}
                                      & \multirow{4}{*}{90}   & mean    & {\cellcolor[rgb]{0.851,0.918,0.827}}46.2 & {\cellcolor[rgb]{0.812,0.886,0.953}}69.9     & {\cellcolor[rgb]{0.851,0.918,0.827}}6.8  & {\cellcolor[rgb]{0.812,0.886,0.953}}13.9     & {\cellcolor[rgb]{0.851,0.918,0.827}}9.9  & {\cellcolor[rgb]{0.812,0.886,0.953}}20.7     & {\cellcolor[rgb]{0.851,0.918,0.827}}23.3 & {\cellcolor[rgb]{0.812,0.886,0.953}}44.9     & {\cellcolor[rgb]{0.851,0.918,0.827}}21.6                                                                                                         & {\cellcolor[rgb]{0.812,0.886,0.953}}37.4                                                                                                               & {\cellcolor[rgb]{0.851,0.918,0.827}}0.49                  & {\cellcolor[rgb]{0.812,0.886,0.953}}0.54                       & {\cellcolor[rgb]{0.851,0.918,0.827}}50.95                 & {\cellcolor[rgb]{0.851,0.918,0.827}}52.35                  \\
                                      &                       & median  & {\cellcolor[rgb]{0.851,0.918,0.827}}45.4 & {\cellcolor[rgb]{0.812,0.886,0.953}}68.8     & {\cellcolor[rgb]{0.851,0.918,0.827}}8.6  & {\cellcolor[rgb]{0.812,0.886,0.953}}22.8     & {\cellcolor[rgb]{0.851,0.918,0.827}}15.8 & {\cellcolor[rgb]{0.812,0.886,0.953}}29.9     & {\cellcolor[rgb]{0.851,0.918,0.827}}25   & {\cellcolor[rgb]{0.812,0.886,0.953}}48.5     & {\cellcolor[rgb]{0.851,0.918,0.827}}23.7                                                                                                         & {\cellcolor[rgb]{0.812,0.886,0.953}}42.5                                                                                                               & {\cellcolor[rgb]{0.851,0.918,0.827}}0.53                  & {\cellcolor[rgb]{0.812,0.886,0.953}}0.61                       & {\cellcolor[rgb]{0.851,0.918,0.827}}45.62                 & {\cellcolor[rgb]{0.851,0.918,0.827}}48.88                  \\
                                      &                       & std     & {\cellcolor[rgb]{0.851,0.918,0.827}}44.8 & {\cellcolor[rgb]{0.812,0.886,0.953}}68.6     & {\cellcolor[rgb]{0.851,0.918,0.827}}13.1 & {\cellcolor[rgb]{0.812,0.886,0.953}}27.6     & {\cellcolor[rgb]{0.851,0.918,0.827}}18.4 & {\cellcolor[rgb]{0.812,0.886,0.953}}33.4     & {\cellcolor[rgb]{0.851,0.918,0.827}}25.7 & {\cellcolor[rgb]{0.812,0.886,0.953}}49.7     & {\cellcolor[rgb]{0.851,0.918,0.827}}25.5                                                                                                         & {\cellcolor[rgb]{0.812,0.886,0.953}}44.8                                                                                                               & {\cellcolor[rgb]{0.851,0.918,0.827}}0.58                  & {\cellcolor[rgb]{0.812,0.886,0.953}}0.64                       & {\cellcolor[rgb]{0.851,0.918,0.827}}40.54                 & {\cellcolor[rgb]{0.851,0.918,0.827}}47.44                  \\
                                      &                       & entropy & {\cellcolor[rgb]{0.851,0.918,0.827}}45.6 & {\cellcolor[rgb]{0.812,0.886,0.953}}67.0     & {\cellcolor[rgb]{0.851,0.918,0.827}}13.9 & {\cellcolor[rgb]{0.812,0.886,0.953}}28.5     & {\cellcolor[rgb]{0.851,0.918,0.827}}19.5 & {\cellcolor[rgb]{0.812,0.886,0.953}}33.8     & {\cellcolor[rgb]{0.851,0.918,0.827}}28.4 & {\cellcolor[rgb]{0.812,0.886,0.953}}53       & {\cellcolor[rgb]{0.851,0.918,0.827}}\textbf{26.8}                                                                                                & {\cellcolor[rgb]{0.812,0.886,0.953}}\textbf{45.6}                                                                                                      & {\cellcolor[rgb]{0.851,0.918,0.827}}\textbf{0.61}         & {\cellcolor[rgb]{0.812,0.886,0.953}}\textbf{0.65}              & {\cellcolor[rgb]{0.851,0.918,0.827}}\textbf{38.43}        & {\cellcolor[rgb]{0.851,0.918,0.827}}\textbf{41.92}         \\ 
\hline
\multirow{8}{*}{Grad}                 & \multirow{2}{*}{25}   & 0.1     & {\cellcolor[rgb]{0.851,0.918,0.827}}44.2 & {\cellcolor[rgb]{0.812,0.886,0.953}}67.8     & {\cellcolor[rgb]{0.851,0.918,0.827}}7.5  & {\cellcolor[rgb]{0.812,0.886,0.953}}16.6     & {\cellcolor[rgb]{0.851,0.918,0.827}}20   & {\cellcolor[rgb]{0.812,0.886,0.953}}34.5     & {\cellcolor[rgb]{0.851,0.918,0.827}}37.2 & {\cellcolor[rgb]{0.812,0.886,0.953}}64.4     & {\cellcolor[rgb]{0.851,0.918,0.827}}\textbf{27.2}                                                                                                & {\cellcolor[rgb]{0.812,0.886,0.953}}\textbf{45.8}                                                                                                      & {\cellcolor[rgb]{0.851,0.918,0.827}}\textbf{0.61}         & {\cellcolor[rgb]{0.812,0.886,0.953}}\textbf{0.66}              & {\cellcolor[rgb]{0.851,0.918,0.827}}\textbf{44.14}        & {\cellcolor[rgb]{0.851,0.918,0.827}}23.93                  \\
                                      &                       & 0.01    & {\cellcolor[rgb]{0.851,0.918,0.827}}29.2 & {\cellcolor[rgb]{0.812,0.886,0.953}}65.7     & {\cellcolor[rgb]{0.851,0.918,0.827}}8.8  & {\cellcolor[rgb]{0.812,0.886,0.953}}18       & {\cellcolor[rgb]{0.851,0.918,0.827}}19.9 & {\cellcolor[rgb]{0.812,0.886,0.953}}34.1     & {\cellcolor[rgb]{0.851,0.918,0.827}}37.9 & {\cellcolor[rgb]{0.812,0.886,0.953}}64.7     & {\cellcolor[rgb]{0.851,0.918,0.827}}24.0                                                                                                         & {\cellcolor[rgb]{0.812,0.886,0.953}}45.6                                                                                                               & {\cellcolor[rgb]{0.851,0.918,0.827}}0.54                  & {\cellcolor[rgb]{0.812,0.886,0.953}}0.65                       & {\cellcolor[rgb]{0.851,0.918,0.827}}54.84                 & {\cellcolor[rgb]{0.851,0.918,0.827}}\textbf{22.49}         \\ 
\hhline{~----------------}
                                      & \multirow{2}{*}{50}   & 0.1     & {\cellcolor[rgb]{0.851,0.918,0.827}}45.7 & {\cellcolor[rgb]{0.812,0.886,0.953}}69.7     & {\cellcolor[rgb]{0.851,0.918,0.827}}9.7  & {\cellcolor[rgb]{0.812,0.886,0.953}}21.4     & {\cellcolor[rgb]{0.851,0.918,0.827}}18.8 & {\cellcolor[rgb]{0.812,0.886,0.953}}32.6     & {\cellcolor[rgb]{0.851,0.918,0.827}}35.2 & {\cellcolor[rgb]{0.812,0.886,0.953}}61.7     & {\cellcolor[rgb]{0.851,0.918,0.827}}27.4                                                                                                         & {\cellcolor[rgb]{0.812,0.886,0.953}}46.4                                                                                                               & {\cellcolor[rgb]{0.851,0.918,0.827}}0.62                  & {\cellcolor[rgb]{0.812,0.886,0.953}}0.67                       & {\cellcolor[rgb]{0.851,0.918,0.827}}42.16                 & {\cellcolor[rgb]{0.851,0.918,0.827}}28.02                  \\
                                      &                       & 0.01    & {\cellcolor[rgb]{0.851,0.918,0.827}}45.4 & {\cellcolor[rgb]{0.812,0.886,0.953}}67.9     & {\cellcolor[rgb]{0.851,0.918,0.827}}11.2 & {\cellcolor[rgb]{0.812,0.886,0.953}}23.1     & {\cellcolor[rgb]{0.851,0.918,0.827}}20   & {\cellcolor[rgb]{0.812,0.886,0.953}}34.9     & {\cellcolor[rgb]{0.851,0.918,0.827}}37.1 & {\cellcolor[rgb]{0.812,0.886,0.953}}64.3     & {\cellcolor[rgb]{0.851,0.918,0.827}}\textbf{28.4}                                                                                                & {\cellcolor[rgb]{0.812,0.886,0.953}}\textbf{47.5}                                                                                                      & {\cellcolor[rgb]{0.851,0.918,0.827}}\textbf{0.64}         & {\cellcolor[rgb]{0.812,0.886,0.953}}\textbf{0.68}              & {\cellcolor[rgb]{0.851,0.918,0.827}}\textbf{40.28}        & {\cellcolor[rgb]{0.851,0.918,0.827}}\textbf{24.13}         \\ 
\hhline{~----------------}
                                      & \multirow{2}{*}{75}   & 0.1     & {\cellcolor[rgb]{0.851,0.918,0.827}}47.5 & {\cellcolor[rgb]{0.812,0.886,0.953}}70.6     & {\cellcolor[rgb]{0.851,0.918,0.827}}9.7  & {\cellcolor[rgb]{0.812,0.886,0.953}}23       & {\cellcolor[rgb]{0.851,0.918,0.827}}18.5 & {\cellcolor[rgb]{0.812,0.886,0.953}}31.6     & {\cellcolor[rgb]{0.851,0.918,0.827}}31.5 & {\cellcolor[rgb]{0.812,0.886,0.953}}57.7     & {\cellcolor[rgb]{0.851,0.918,0.827}}26.8                                                                                                         & {\cellcolor[rgb]{0.812,0.886,0.953}}45.7                                                                                                               & {\cellcolor[rgb]{0.851,0.918,0.827}}0.61                  & {\cellcolor[rgb]{0.812,0.886,0.953}}0.66                       & {\cellcolor[rgb]{0.851,0.918,0.827}}40.97                 & {\cellcolor[rgb]{0.851,0.918,0.827}}35.58                  \\
                                      &                       & 0.01    & {\cellcolor[rgb]{0.851,0.918,0.827}}47.0 & {\cellcolor[rgb]{0.812,0.886,0.953}}71.6     & {\cellcolor[rgb]{0.851,0.918,0.827}}21.1 & {\cellcolor[rgb]{0.812,0.886,0.953}}36.5     & {\cellcolor[rgb]{0.851,0.918,0.827}}19.2 & {\cellcolor[rgb]{0.812,0.886,0.953}}32.6     & {\cellcolor[rgb]{0.851,0.918,0.827}}32.3 & {\cellcolor[rgb]{0.812,0.886,0.953}}59.4     & {\cellcolor[rgb]{0.851,0.918,0.827}}\textbf{29.9}                                                                                                & {\cellcolor[rgb]{0.812,0.886,0.953}}\textbf{50.0}                                                                                                      & {\cellcolor[rgb]{0.851,0.918,0.827}}\textbf{0.67}         & {\cellcolor[rgb]{0.812,0.886,0.953}}\textbf{0.72}              & {\cellcolor[rgb]{0.851,0.918,0.827}}\textbf{31.96}        & {\cellcolor[rgb]{0.851,0.918,0.827}}\textbf{33.95}         \\ 
\hhline{~----------------}
                                      & \multirow{2}{*}{90}   & 0.1     & {\cellcolor[rgb]{0.851,0.918,0.827}}48.7 & {\cellcolor[rgb]{0.812,0.886,0.953}}72.9     & {\cellcolor[rgb]{0.851,0.918,0.827}}15.6 & {\cellcolor[rgb]{0.812,0.886,0.953}}31.1     & {\cellcolor[rgb]{0.851,0.918,0.827}}17.7 & {\cellcolor[rgb]{0.812,0.886,0.953}}32       & {\cellcolor[rgb]{0.851,0.918,0.827}}28   & {\cellcolor[rgb]{0.812,0.886,0.953}}53.1     & {\cellcolor[rgb]{0.851,0.918,0.827}}27.5                                                                                                         & {\cellcolor[rgb]{0.812,0.886,0.953}}47.3                                                                                                               & {\cellcolor[rgb]{0.851,0.918,0.827}}0.62                  & {\cellcolor[rgb]{0.812,0.886,0.953}}0.68                       & {\cellcolor[rgb]{0.851,0.918,0.827}}36.09                 & {\cellcolor[rgb]{0.851,0.918,0.827}}\textbf{42.74}         \\
                                      &                       & 0.01    & {\cellcolor[rgb]{0.851,0.918,0.827}}49.2 & {\cellcolor[rgb]{0.812,0.886,0.953}}73.5     & {\cellcolor[rgb]{0.851,0.918,0.827}}20.4 & {\cellcolor[rgb]{0.812,0.886,0.953}}39.4     & {\cellcolor[rgb]{0.851,0.918,0.827}}18   & {\cellcolor[rgb]{0.812,0.886,0.953}}32.3     & {\cellcolor[rgb]{0.851,0.918,0.827}}27.9 & {\cellcolor[rgb]{0.812,0.886,0.953}}53.7     & {\cellcolor[rgb]{0.851,0.918,0.827}}\textbf{28.9}                                                                                                & {\cellcolor[rgb]{0.812,0.886,0.953}}\textbf{49.7}                                                                                                      & {\cellcolor[rgb]{0.851,0.918,0.827}}\textbf{0.65}         & {\cellcolor[rgb]{0.812,0.886,0.953}}\textbf{0.71}              & {\cellcolor[rgb]{0.851,0.918,0.827}}\textbf{31.69}        & {\cellcolor[rgb]{0.851,0.918,0.827}}42.94                  \\ 
\hline
\multirow{4}{*}{MMN}                  & 25                    & -       & {\cellcolor[rgb]{0.851,0.918,0.827}}44.6 & {\cellcolor[rgb]{0.812,0.886,0.953}}68.0     & {\cellcolor[rgb]{0.851,0.918,0.827}}5.1  & {\cellcolor[rgb]{0.812,0.886,0.953}}12.2     & {\cellcolor[rgb]{0.851,0.918,0.827}}17.8 & {\cellcolor[rgb]{0.812,0.886,0.953}}31.3     & {\cellcolor[rgb]{0.851,0.918,0.827}}33.5 & {\cellcolor[rgb]{0.812,0.886,0.953}}60       & {\cellcolor[rgb]{0.851,0.918,0.827}}25.3                                                                                                         & {\cellcolor[rgb]{0.812,0.886,0.953}}42.9                                                                                                               & {\cellcolor[rgb]{0.851,0.918,0.827}}0.57                  & {\cellcolor[rgb]{0.812,0.886,0.953}}0.62                       & {\cellcolor[rgb]{0.851,0.918,0.827}}47.36                 & {\cellcolor[rgb]{0.851,0.918,0.827}}31.49                  \\
                                      & 50                    & -       & {\cellcolor[rgb]{0.851,0.918,0.827}}47.3 & {\cellcolor[rgb]{0.812,0.886,0.953}}69.7     & {\cellcolor[rgb]{0.851,0.918,0.827}}4.2  & {\cellcolor[rgb]{0.812,0.886,0.953}}10.1     & {\cellcolor[rgb]{0.851,0.918,0.827}}17.4 & {\cellcolor[rgb]{0.812,0.886,0.953}}31.7     & {\cellcolor[rgb]{0.851,0.918,0.827}}31.5 & {\cellcolor[rgb]{0.812,0.886,0.953}}58       & {\cellcolor[rgb]{0.851,0.918,0.827}}25.1                                                                                                         & {\cellcolor[rgb]{0.812,0.886,0.953}}42.4                                                                                                               & {\cellcolor[rgb]{0.851,0.918,0.827}}0.57                  & {\cellcolor[rgb]{0.812,0.886,0.953}}0.61                       & {\cellcolor[rgb]{0.851,0.918,0.827}}46.33                 & {\cellcolor[rgb]{0.851,0.918,0.827}}35.58                  \\
                                      & 75                    & -       & {\cellcolor[rgb]{0.851,0.918,0.827}}49.4 & {\cellcolor[rgb]{0.812,0.886,0.953}}72.7     & {\cellcolor[rgb]{0.851,0.918,0.827}}6.7  & {\cellcolor[rgb]{0.812,0.886,0.953}}15.9     & {\cellcolor[rgb]{0.851,0.918,0.827}}15.5 & {\cellcolor[rgb]{0.812,0.886,0.953}}28.8     & {\cellcolor[rgb]{0.851,0.918,0.827}}28.1 & {\cellcolor[rgb]{0.812,0.886,0.953}}52.1     & {\cellcolor[rgb]{0.851,0.918,0.827}}24.9                                                                                                         & {\cellcolor[rgb]{0.812,0.886,0.953}}42.4                                                                                                               & {\cellcolor[rgb]{0.851,0.918,0.827}}0.56                  & {\cellcolor[rgb]{0.812,0.886,0.953}}0.61                       & {\cellcolor[rgb]{0.851,0.918,0.827}}44.16                 & {\cellcolor[rgb]{0.851,0.918,0.827}}42.54                  \\
                                      & 90                    & -       & {\cellcolor[rgb]{0.851,0.918,0.827}}48.6 & {\cellcolor[rgb]{0.812,0.886,0.953}}72.0     & {\cellcolor[rgb]{0.851,0.918,0.827}}10.4 & {\cellcolor[rgb]{0.812,0.886,0.953}}18.6     & {\cellcolor[rgb]{0.851,0.918,0.827}}14.2 & {\cellcolor[rgb]{0.812,0.886,0.953}}26.8     & {\cellcolor[rgb]{0.851,0.918,0.827}}13.8 & {\cellcolor[rgb]{0.812,0.886,0.953}}32.5     & {\cellcolor[rgb]{0.851,0.918,0.827}}21.7                                                                                                         & {\cellcolor[rgb]{0.812,0.886,0.953}}37.5                                                                                                               & {\cellcolor[rgb]{0.851,0.918,0.827}}0.49                  & {\cellcolor[rgb]{0.812,0.886,0.953}}0.54                       & {\cellcolor[rgb]{0.851,0.918,0.827}}42.97                 & {\cellcolor[rgb]{0.851,0.918,0.827}}71.78                  \\ 
\hline
\multicolumn{1}{l}{Fine tuning}       & \multicolumn{1}{r}{-} & -       & {\cellcolor[rgb]{0.851,0.918,0.827}}44.2 & {\cellcolor[rgb]{0.812,0.886,0.953}}66.6     & {\cellcolor[rgb]{0.851,0.918,0.827}}5.4  & {\cellcolor[rgb]{0.812,0.886,0.953}}12.8     & {\cellcolor[rgb]{0.851,0.918,0.827}}12   & {\cellcolor[rgb]{0.812,0.886,0.953}}23.5     & {\cellcolor[rgb]{0.851,0.918,0.827}}34.9 & {\cellcolor[rgb]{0.812,0.886,0.953}}61.5     & {\cellcolor[rgb]{0.851,0.918,0.827}}24.1                                                                                                         & {\cellcolor[rgb]{0.812,0.886,0.953}}41.1                                                                                                               & {\cellcolor[rgb]{0.851,0.918,0.827}}0.54                  & {\cellcolor[rgb]{0.812,0.886,0.953}}0.59                       & {\cellcolor[rgb]{0.851,0.918,0.827}}52.02                 & {\cellcolor[rgb]{0.851,0.918,0.827}}28.63                  \\ 
\hline
\multicolumn{1}{l}{Experience Replay} & \multicolumn{1}{r}{-} & -       & {\cellcolor[rgb]{0.851,0.918,0.827}}46.7 & {\cellcolor[rgb]{0.812,0.886,0.953}}71.3     & {\cellcolor[rgb]{0.851,0.918,0.827}}21.5 & {\cellcolor[rgb]{0.812,0.886,0.953}}37.8     & {\cellcolor[rgb]{0.851,0.918,0.827}}24.9 & {\cellcolor[rgb]{0.812,0.886,0.953}}40.6     & {\cellcolor[rgb]{0.851,0.918,0.827}}42.5 & {\cellcolor[rgb]{0.812,0.886,0.953}}71.9     & {\cellcolor[rgb]{0.851,0.918,0.827}}33.9                                                                                                         & {\cellcolor[rgb]{0.812,0.886,0.953}}55.4                                                                                                               & {\cellcolor[rgb]{0.851,0.918,0.827}}0.77                  & {\cellcolor[rgb]{0.812,0.886,0.953}}0.80                       & {\cellcolor[rgb]{0.851,0.918,0.827}}27.40                 & {\cellcolor[rgb]{0.851,0.918,0.827}}13.09                  \\ 
\hline
\multicolumn{1}{l}{Ground Truth}      & \multicolumn{1}{r}{-} & -       & {\cellcolor[rgb]{0.851,0.918,0.827}}56.8 & {\cellcolor[rgb]{0.812,0.886,0.953}}83.2     & {\cellcolor[rgb]{0.851,0.918,0.827}}35.7 & {\cellcolor[rgb]{0.812,0.886,0.953}}58.1     & {\cellcolor[rgb]{0.851,0.918,0.827}}35.8 & {\cellcolor[rgb]{0.812,0.886,0.953}}62.1     & {\cellcolor[rgb]{0.851,0.918,0.827}}48.9 & {\cellcolor[rgb]{0.812,0.886,0.953}}75.3     & {\cellcolor[rgb]{0.851,0.918,0.827}}44.3                                                                                                         & {\cellcolor[rgb]{0.812,0.886,0.953}}69.7                                                                                                               & {\cellcolor[rgb]{0.851,0.918,0.827}}-                     & {\cellcolor[rgb]{0.812,0.886,0.953}}-                          & {\cellcolor[rgb]{0.851,0.918,0.827}}-                     & {\cellcolor[rgb]{0.851,0.918,0.827}}-                      \\
\hline
\end{tabular}}
\end{table*}

Table \ref{tab:taesa-results} summarizes the results on the proposed benchmark with the green color highlighting metrics related to $mAP$ and blue for $mAP_{[.50]}$. As the benchmark involves class-incremental and domain-incremental aspects, we noticed that when there is little drift in the appearance of previously known objects that show up in the new task images, these instances reinforce the ``old knowledge'' and can be considered as a small case of replay. This can be checked by the fact that the forgetting in the fine-tuning approach is ``soft'' when compared to other artificial benchmarks, such as Incremental Pascal VOC, in which classes that do not appear in further training sets are completely forgotten. Furthermore, the benchmark was organized in a way that minimized label conflicts, leading to less interference in the weights assigned to each class.

Applying a penalty to the gradients of important parameters improved the results of leaving them frozen (i.e. MMN) in all scenarios. The best results were seen when applying a 1\% of the penalty to 50\% or more of the important weights. Due to a slight imbalance between the number of available data and classes in each task and the fact that the first task had more learning steps, it was found that keeping most of the old weights unchanged, or slightly adjusting them to new tasks, proved to be effective for average performance. However, when checking the performance in the intermediate tasks (i.e., Tasks 2 and 3) and comparing them to the fine-tuning and upper-bound results, we see that forgetting still occurs, but to a lesser extent than in the other evaluated methods.  

Selecting the most important layers based on information entropy was the most impartial in terms of the percentage of layers chosen, and generally yielded superior outcomes compared to other statistical measures. Yet, freezing 75\% of the layers based on the mean of feature map activations seemed to produce the best results, achieving a good balance in the final $\Omega_{mAP}$ and $\Omega_{mAP[.50]}$, although it significantly impacted knowledge retention in intermediate tasks The other layer-freezing methods attained similar results, but with less forgetting in the intermediate tasks. This highlights the necessity to look at the big picture and not only specific metrics based on averages.

Although the full benchmark seemed challenging by having to deal with new classes and domains, the initial task's diverse and abundant data helped prepare the model to learn with small adjustments in new task scenarios. All evaluated strategies performed better than fine-tuning and MMN baselines but fell behind the results achieved through experience replay. For scenarios where saving samples is not feasible, a hybrid strategy involving parameter isolation and fake labeling may help reduce the gap in performance against replay methods. Nevertheless, when possible, combining these methods with parameter-isolation strategies can be seen as a promising direction for investigation.

\section{\uppercase{Conclusions}}
\label{sec:conclusion}

In this paper, we discussed different ways to mitigate forgetting when learning new object detection tasks by using simple criteria to freeze layers and heuristics for how important parameters should be updated. We found that mining and freezing layers based on feature map statistics, particularly on their information entropy, yielded better results than freezing individual neurons when updating the network with data from a single class. However, when introducing data from several classes, the simple arrangements brought by the layer-freezing strategy were not as successful. The layer-freezing strategies mostly outperformed the mining of individual neurons but presented lower performance when directly compared to more traditional and complex knowledge-distillation methods such as ILOD and RILOD, or experience replay. Additionally, results also showed that applying individual penalties to the gradients of important neurons did not significantly differ from the possibility of freezing them.

As a future line of work, it may be beneficial to explore fine-grained freezing solutions that involve mining and freezing individual convolutional filters based on their internal statistics. Hybrid techniques that balance learning with the use of experience replay could also be proposed to prevent forgetting and adapt more quickly to new scenarios. Furthermore, it would be useful to investigate measures of task-relatedness as a means of defining the freezing coefficients among sequential updates.

\section*{\uppercase{Acknowledgements}}

This study was funded in part by the Coordenação de Aperfeiçoamento de Pessoal de Nível Superior - Brasil (CAPES) - Finance Code 001 and by ANEEL (\textit{Agência Nacional de Energia Elétrica}) and TAESA (Transmissora Aliança de Energia Elétrica S.A.), project PD-07130-0059/2020. The authors also would like to thank the Conselho Nacional de Desenvolvimento Científico e Tecnológico (CNPq) and the Eldorado Research Institute for supporting this research.

\bibliographystyle{apalike.bst}
{\small
\bibliography{example.bib}}

\begin{thebibliography}{}

\bibitem[Chaudhry et~al., 2018]{chaudhry2018efficient}
Chaudhry, A., Ranzato, M., Rohrbach, M., and Elhoseiny, M. (2018).
\newblock Efficient lifelong learning with a-gem.
\newblock {\em arXiv preprint arXiv:1812.00420}.

\bibitem[Chen et~al., 2019]{chen2019mmdetection}
Chen, K., Wang, J., Pang, J., Cao, Y., Xiong, Y., Li, X., Sun, S., Feng, W., Liu, Z., Xu, J., et~al. (2019).
\newblock Mmdetection: Open mmlab detection toolbox and benchmark.
\newblock {\em arXiv preprint arXiv:1906.07155}.

\bibitem[Delange et~al., 2021]{delange2021continual}
Delange, M., Aljundi, R., Masana, M., Parisot, S., Jia, X., Leonardis, A., Slabaugh, G., and Tuytelaars, T. (2021).
\newblock A continual learning survey: Defying forgetting in classification tasks.
\newblock {\em IEEE Transactions on Pattern Analysis and Machine Intelligence}.

\bibitem[Hadsell et~al., 2020]{hadsell2020embracing}
Hadsell, R., Rao, D., Rusu, A.~A., and Pascanu, R. (2020).
\newblock Embracing change: Continual learning in deep neural networks.
\newblock {\em Trends in cognitive sciences}.

\bibitem[Kirkpatrick et~al., 2017]{kirkpatrick2017overcoming}
Kirkpatrick, J., Pascanu, R., Rabinowitz, N., Veness, J., Desjardins, G., Rusu, A.~A., Milan, K., Quan, J., Ramalho, T., Grabska-Barwinska, A., et~al. (2017).
\newblock Overcoming catastrophic forgetting in neural networks.
\newblock {\em Proceedings of the national academy of sciences}, 114(13):3521--3526.

\bibitem[LeCun et~al., 1989]{lecun1989optimal}
LeCun, Y., Denker, J., and Solla, S. (1989).
\newblock Optimal brain damage.
\newblock {\em Advances in neural information processing systems}, 2.

\bibitem[Li et~al., 2019]{li2019rilod}
Li, D., Tasci, S., Ghosh, S., Zhu, J., Zhang, J., and Heck, L. (2019).
\newblock Rilod: Near real-time incremental learning for object detection at the edge.
\newblock In {\em Proceedings of the 4th ACM/IEEE Symposium on Edge Computing}, pages 113--126.

\bibitem[Li et~al., 2016]{li2016pruning}
Li, H., Kadav, A., Durdanovic, I., Samet, H., and Graf, H.~P. (2016).
\newblock Pruning filters for efficient convnets.
\newblock {\em arXiv preprint arXiv:1608.08710}.

\bibitem[Li et~al., 2018]{li2018incremental}
Li, W., Wu, Q., Xu, L., and Shang, C. (2018).
\newblock Incremental learning of single-stage detectors with mining memory neurons.
\newblock In {\em 2018 IEEE 4th International Conference on Computer and Communications (ICCC)}, pages 1981--1985. IEEE.

\bibitem[Li and Hoiem, 2017]{li2017learning}
Li, Z. and Hoiem, D. (2017).
\newblock Learning without forgetting.
\newblock {\em IEEE transactions on pattern analysis and machine intelligence}, 40(12):2935--2947.

\bibitem[Lin et~al., 2017]{lin2017focal}
Lin, T.-Y., Goyal, P., Girshick, R., He, K., and Doll{\'a}r, P. (2017).
\newblock Focal loss for dense object detection.
\newblock In {\em Proceedings of the IEEE international conference on computer vision}, pages 2980--2988.

\bibitem[Liu and Wu, 2019]{liu2019channel}
Liu, C. and Wu, H. (2019).
\newblock Channel pruning based on mean gradient for accelerating convolutional neural networks.
\newblock {\em Signal Processing}, 156:84--91.

\bibitem[Luo and Wu, 2017]{luo2017entropy}
Luo, J.-H. and Wu, J. (2017).
\newblock An entropy-based pruning method for cnn compression.
\newblock {\em arXiv preprint arXiv:1706.05791}.

\bibitem[Mallya and Lazebnik, 2018]{mallya2018packnet}
Mallya, A. and Lazebnik, S. (2018).
\newblock Packnet: Adding multiple tasks to a single network by iterative pruning.
\newblock In {\em Proceedings of the IEEE conference on Computer Vision and Pattern Recognition}, pages 7765--7773.

\bibitem[Menezes et~al., 2023]{menezes2022continual}
Menezes, A.~G., de~Moura, G., Alves, C., and de~Carvalho, A.~C. (2023).
\newblock Continual object detection: A review of definitions, strategies, and challenges.
\newblock {\em Neural Networks}.

\bibitem[Mirzadeh et~al., 2021]{mirzadeh2021wide}
Mirzadeh, S.~I., Chaudhry, A., Hu, H., Pascanu, R., Gorur, D., and Farajtabar, M. (2021).
\newblock Wide neural networks forget less catastrophically.
\newblock {\em arXiv preprint arXiv:2110.11526}.

\bibitem[Shaheen et~al., 2021]{shaheen2021continual}
Shaheen, K., Hanif, M.~A., Hasan, O., and Shafique, M. (2021).
\newblock Continual learning for real-world autonomous systems: Algorithms, challenges and frameworks.
\newblock {\em arXiv preprint arXiv:2105.12374}.

\bibitem[Shmelkov et~al., 2017]{shmelkov2017incremental}
Shmelkov, K., Schmid, C., and Alahari, K. (2017).
\newblock Incremental learning of object detectors without catastrophic forgetting.
\newblock In {\em Proceedings of the IEEE international conference on computer vision}, pages 3400--3409.

\bibitem[Tian et~al., 2020]{tian2020fcos}
Tian, Z., Shen, C., Chen, H., and He, T. (2020).
\newblock Fcos: A simple and strong anchor-free object detector.
\newblock {\em IEEE Transactions on Pattern Analysis and Machine Intelligence}.

\bibitem[ul~Haq et~al., 2021]{ul2021incremental}
ul~Haq, Q.~M., Ruan, S.-J., Haq, M.~A., Karam, S., Shieh, J.~L., Chondro, P., and Gao, D.-Q. (2021).
\newblock An incremental learning of yolov3 without catastrophic forgetting for smart city applications.
\newblock {\em IEEE Consumer Electronics Magazine}.

\bibitem[Wang et~al., 2021]{wang2021filter}
Wang, J., Jiang, T., Cui, Z., and Cao, Z. (2021).
\newblock Filter pruning with a feature map entropy importance criterion for convolution neural networks compressing.
\newblock {\em Neurocomputing}, 461:41--54.

\bibitem[Wu et~al., 2020]{wu2020recent}
Wu, X., Sahoo, D., and Hoi, S.~C. (2020).
\newblock Recent advances in deep learning for object detection.
\newblock {\em Neurocomputing}, 396:39--64.

\bibitem[Zenke et~al., 2017]{zenke2017continual}
Zenke, F., Poole, B., and Ganguli, S. (2017).
\newblock Continual learning through synaptic intelligence.
\newblock In {\em International conference on machine learning}, pages 3987--3995. PMLR.

\bibitem[Zou et~al., 2019]{zou2019object}
Zou, Z., Shi, Z., Guo, Y., and Ye, J. (2019).
\newblock Object detection in 20 years: A survey.
\newblock {\em arXiv preprint arXiv:1905.05055}.

\end{thebibliography}


\end{document}